\documentclass[sn-basic, Numbered]{sn-jnl}

\usepackage{ragged2e,booktabs,tabularx}
\usepackage{graphicx}%
\usepackage{multirow}%
\usepackage{amsmath,amssymb,amsfonts}%
\usepackage{amsthm}%
\usepackage{mathrsfs}%
\usepackage[title]{appendix}%
\usepackage{xcolor}%
\usepackage{textcomp}%
\usepackage{manyfoot}%
\usepackage{booktabs}%
\usepackage{algorithm}%
\usepackage{algorithmicx}%
\usepackage{algpseudocode}%
\usepackage{listings}%
\usepackage{hyperref}
\usepackage{adjustbox}
\usepackage{bbding}
\usepackage{arydshln}
\usepackage{anyfontsize}

\begin{document}

\title[Article Title]{AD-Net: Attention-based dilated convolutional residual network with guided decoder for robust skin lesion segmentation}


\author*[1,2]{\fnm{Asim} \sur{Naveed} }\email{asimnaveed@uet.edu.pk}

\author[1]{\fnm{Syed S.} \sur{Naqvi}}\email{saud\_naqvi@comsats.edu.pk}

\author[4]{\fnm{Tariq M.} \sur{Khan}}\email{tariq045@gmail.com}

\author[1]{\fnm{Shahzaib} \sur{Iqbal}}\email{shahzaib.iqbal91@gmail.com}

\author[3]{\fnm{M. Yaqoob} \sur{Wani}}\email{muhammad.yaqoob@cs.iiui.edu.pk}

\author[1]{\fnm{Haroon Ahmed  \sur{Khan}}}\email{haroon.ahmed@comsats.edu.pk}

\affil*[1]{\orgdiv{Department of Electrical and Computer Engineering}, \orgname{COMSATS University Islamabad (CUI)}, \city{Islamabad}, \postcode{45550}, \country{Pakistan}}
\affil*[2]{\orgdiv{Department of Computer Science and Engineering}, \orgname{University of Engineering and Technology (UET) Lahore}, \city{Narowal Campus}, \postcode{51600}, \country{Pakistan}}
\affil[3]{\orgdiv{Department of Computer Science \& Information Technology}, \orgname{{IBADAT International University Islamabad}, \city{Islamabad}, \postcode{46000}, \country{Pakistan}}}
\affil[4]{\orgdiv{School of Computer Science and Engineering}, \orgname{{ University of New South Wales}, \city{Sydney},  \postcode{1466}, \country{Australia}}
}


\abstract {In computer-aided diagnosis tools employed for skin cancer treatment and early diagnosis, skin lesion segmentation is important. However, achieving precise segmentation is challenging due to inherent variations in appearance, contrast, texture, and blurry lesion boundaries. This research presents a robust approach utilizing a dilated convolutional residual network, which incorporates an attention-based spatial feature enhancement block (ASFEB) and employs a guided decoder strategy. In each dilated convolutional residual block, dilated convolution is employed to broaden the receptive field with varying dilation rates. To improve the spatial feature information of the encoder, we employed an attention-based spatial feature enhancement block in the skip connections. The ASFEB in our proposed method combines feature maps obtained from average and maximum-pooling operations. These combined features are then weighted using the active outcome of global average pooling and convolution operations. Additionally, we have incorporated a guided decoder strategy, where each decoder block is optimized using an individual loss function to enhance the feature learning process in the proposed AD-Net. The proposed AD-Net presents a significant benefit by necessitating fewer model parameters compared to its peer methods. This reduction in parameters directly impacts the number of labeled data required for training, facilitating faster convergence during the training process. The effectiveness of the proposed AD-Net was evaluated using four public benchmark datasets. We conducted a Wilcoxon signed-rank test to verify the efficiency of the AD-Net. The outcomes suggest that our method surpasses other cutting-edge methods in performance, even without the implementation of data augmentation strategies.}


\maketitle

\section{Introduction}\label{sec1}
\begin{figure*}[htbp]
    \centering
    \includegraphics[width = \textwidth]{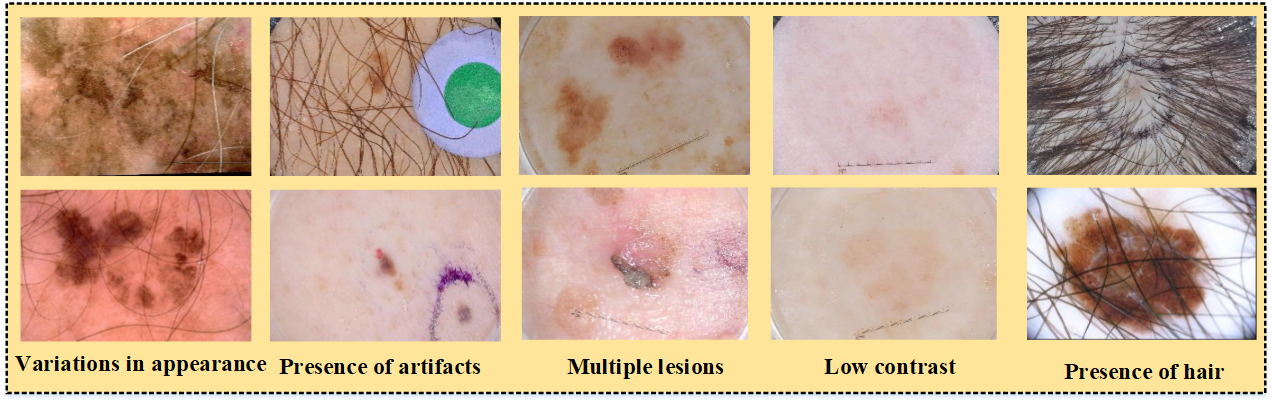}
    \caption{{Dermoscopic skin lesion images present several challenges}}
    \label{Frameworkchallenges}
\end{figure*}

Skin diseases indeed represent a substantial health concern, with skin cancer, particularly melanoma, being one of the most life-threatening forms. According to global cancer statistics, skin cancers are among the fastest-growing cancers worldwide \cite{siegel2023cancer}. 
The American Cancer Society (ACS) reports that cancer is the leading cause of mortality globally, accounting for 1.96 million new cancer cases in 2023, and 0.61 million deaths due to cancer. In 2023, 0.098 million new cases of skin cancer from melanoma were reported, with 8. 2\% of individuals losing their lives \cite{siegel2023cancer}.
Skin cancer is commonly classified into two main types: melanoma and non-melanoma \cite{balch2009final}. Non-melanoma skin cancers, including squamous cell carcinoma and basal cell carcinoma, are generally considered less life-threatening. However, they may still require painful treatment interventions. In contrast, melanoma represents a highly malignant and deadly form of skin cancer, characterised by a significantly higher mortality rate \cite{zhang2023accpg}.
In the realm of skin cancer management, prompt diagnosis and timely treatment are important factors for effective control. However, relying solely on visual assessments by medical experts can introduce subjectivity and lead to inconsistent diagnoses, even among experienced professionals. Thus, to enable an early and precise diagnosis, the establishment of effective and automated techniques for the segmentation of skin lesions is essential.
These automated methods offer consistent and objective analyses, which improves the reliability and efficiency of skin cancer diagnosis and treatment \cite{maqsood2023multiclass, sufyan2023artificial, litjens2017survey}.
Accurate segmentation of skin lesions is a critical prerequisite for effective diagnosis, analysis, and treatment in computer-aided diagnostic (CAD) systems. However, segmentation of dermoscopic images poses distinct challenges, primarily due to variations in colour and texture, as well as the presence of artefacts such as hair and marks \cite{DAI2022102293}. Dermoscopy is a noninvasive imaging technique that enables in vivo observation of pigmented skin lesions and employs optical magnification lenses and specialised illumination to improve the visibility of the underlying features \cite{panayides2020ai}. Dermoscopic images of skin lesions pose further challenges. First, they include the irregular and fuzzy boundaries typically associated with skin lesions. Secondly, distinguishing a skin lesion from its surrounding tissue is often difficult. Third, interpreting the features of skin lesions can be challenging due to their typically irregular shapes and colors. In addition, segmentation is complicated by various interference factors, such as hairs, blood vessels, ruler markings, and ink speckles \cite{gu2022net, FENG2022105942}. The challenges mentioned above are presented in Figure~\ref{Frameworkchallenges}. 

Traditional image segmentation methods often depend on manually crafted features \cite{khan2016stopping, soomro2016automatic,khan2019boosting,naqvi2019automatic}, which show limited performance when it comes to segmenting complex images such as dermoscopic images\cite{abdullah2021review,imtiaz2021screening,khan2022neural}. Manually crafted features require domain experience and may not generalise well to the wide variability in the appearance of the lesion \cite{7004778}. In fact, deep learning techniques have revolutionised the domain by their ability to learn directly from data, yielding remarkable results \cite{khan2020exploiting,khan2020semantically,khan2023retinal}. Unlike traditional algorithms that often rely on hand-crafted features, deep learning models operate on a data-driven basis, allowing them to automatically extract relevant features and patterns from input data \cite{khan2021residual,khan2022t,iqbal2022g,arsalan2022prompt,khan2022width}. This data-driven approach contributes to the robustness of segmentation models and builds confidence in their performance \cite{khan2022mkis, naqvi2023glan, khan2023feature,khan2024esdmr,khan2024lmbf,javed2024region}. In clinical settings, this confidence is crucial, as it serves as an acceptance criterion for the deployment of such models in real-world settings \cite{holzinger2022next,iqbal2024tesl,farooq2024lssf,iqbal2024tbconvl}. 

The U-Net architecture has gathered substantial popularity in the realm of medical image segmentation, due to its remarkable performance in capturing fine details of features via its encoder-decoder paths with skip connections \cite{ronneberger2015u}. Based on U-Net, researchers have developed several unique architectures, such as U-Net++ \cite{zhou2018unet++}, Attention U-Net \cite{oktay2018attention}, and recurrent residual U-Net \cite{alom2019recurrent}, which are specifically designed for different segmentation tasks. In recent years, various advanced techniques and modifications based on the U-Net framework have emerged to enhance both performance and computational efficiency in tasks such as segmentation of skin lesions \cite{wu2020automated}.  
\begin{figure*}[htbp]
    \centering
    \includegraphics[scale=0.73]{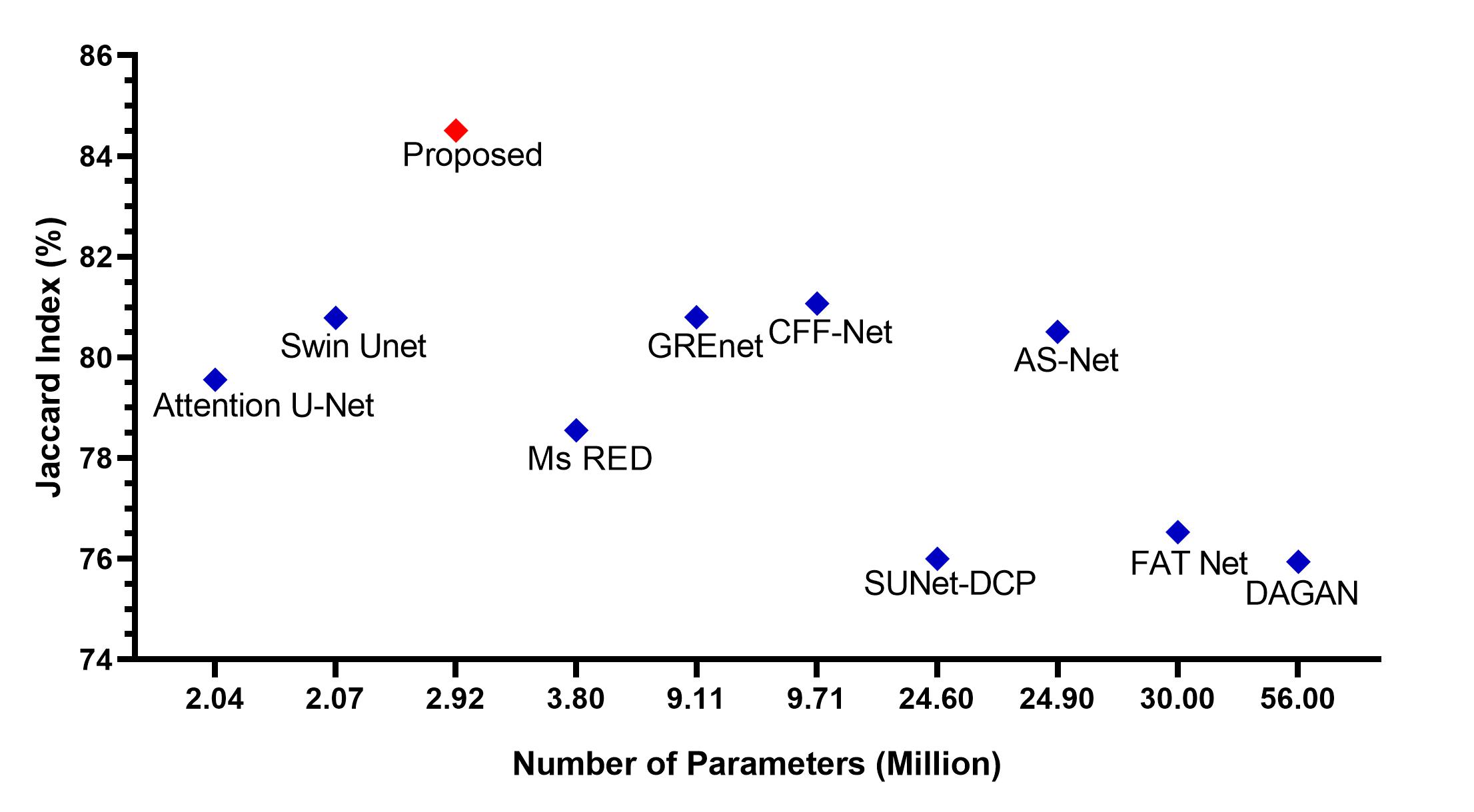}
    \caption{Parameters vs Jaccard Index details}
    \label{parameters}
\end{figure*}

Many researchers proposed methods for the segmentation of skin lesions with different approaches. For instance, Lei \textit{et~al.} introduced an approach called the Dual Adversarial Generator and Discriminator Network (DAGAN),  which employs dual discriminators to analyse the boundaries of objects and contextual relationships \cite{LEI2020101716}. To achieve better performance, the AS-Net authors introduced a network that blends spatial attention with channel attention techniques \cite{HU2022117112}. Ms RED is an attention-based multiscale feature fusion method that improved the segmentation efficiency of their method by incorporating different components into the proposed approach \cite{ DAI2022102293}. FAT-Net, a transformer-based encoder-decoder approach, achieved performance on the skin lesion segmentation task \cite{WU2022102327}. Some more recent methods such as GREnet \cite{wang2023grenet}, CFF-Net \cite{qin2023dynamically}, and SUNet-DCP \cite{song2023decoupling} also obtained competent results for segmentation of skin lesions. Most of these methods are heavy parameter methods and their performance is limited due to the learning of redundant features. 
Figure~\ref{parameters} illustrates the performance of the proposed AD-Net compared to state-of-the-art (SOTA) methods on the ISIC 2017 dataset. The comparison of the Jaccard index values among various SOTA methods, correlated with the number of trainable parameters, is presented. 

\begin{figure*}[htbp]
    \centering
    \includegraphics[scale=0.24]{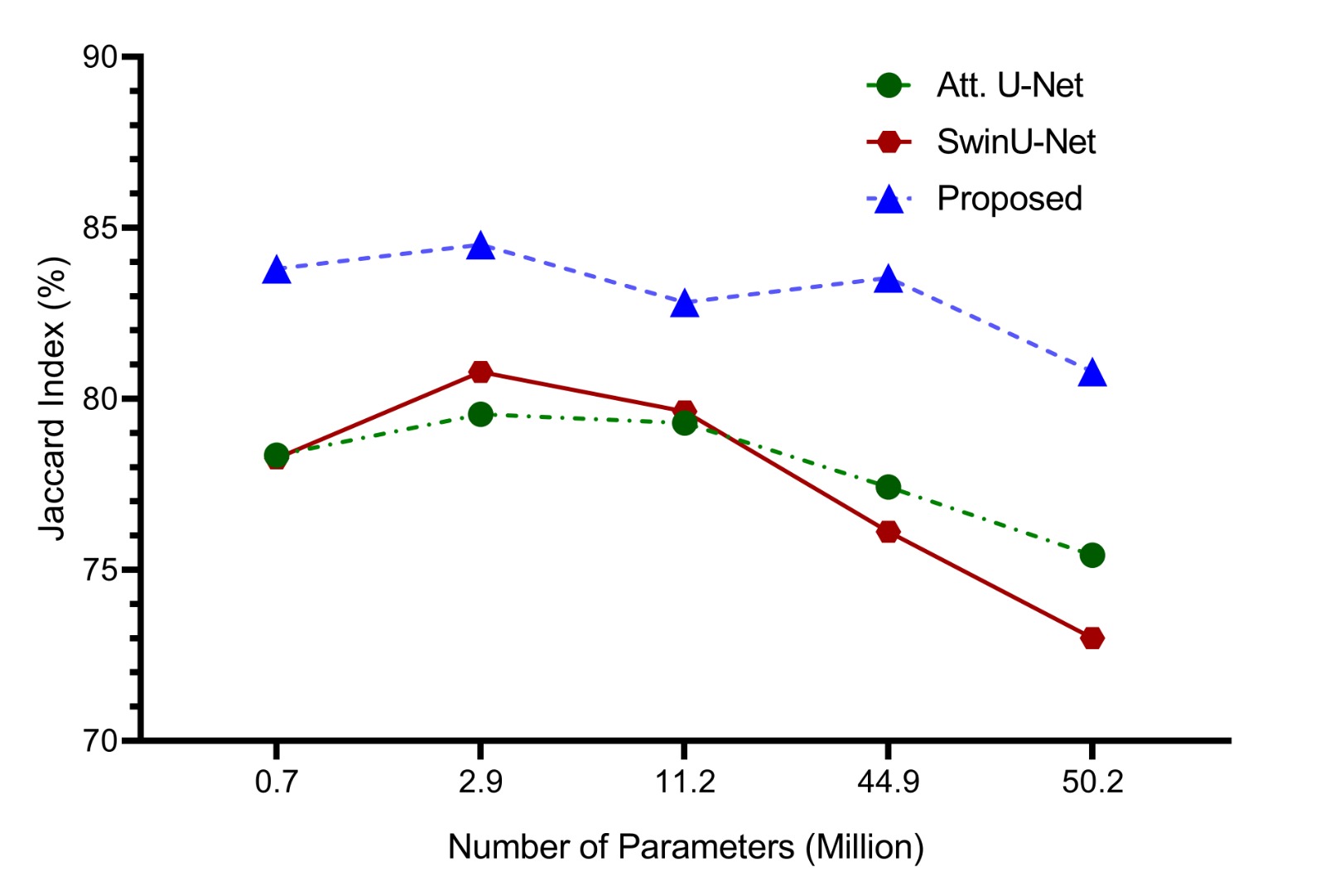}
    \caption {{Performance of proposed method, Transformer model \cite{cao2023swin} and attention model \cite{oktay2018attention} with respect to parameters}}
    \label{parameterspsa}
\end{figure*}

To observe the impact of trainable parameters on model performance, we conducted an evaluation using the transformer-based method \cite{cao2023swin} and the attention-based method \cite{oktay2018attention}. Specifically, we assessed how the variation of the number of parameters influences the Jaccard index value, a key metric to measure segmentation quality. The results are illustrated in Figure~\ref{parameterspsa}.
Our findings indicate that an increase in the number of trainable parameters does not necessarily result in better performance. In contrast, performance tends to decline as the number of parameters increases, likely due to redundant features in the model learning. This evaluation suggests that an excessive number of parameters can lead to overfitting, where the model captures noise rather than relevant patterns, thereby diminishing its generalization ability.
Figure~\ref{parameterspsa} clearly shows this trend, highlighting the inverse relationship between the parameter count and the Jaccard index beyond a certain point. This comparison is crucial for optimising model configurations in skin lesion segmentation tasks. By carefully analyzing these outcomes, we can determine the optimal number of parameters that maximize performance without incurring the drawbacks of overfitting. This balance is essential for developing robust and efficient models that deliver high performance in clinical applications.

Khan \textit{et~al.} suggested a criterion for the choice of the optimal model regarding the complexity of the image. This study provides insight and guidelines for obtaining effective segmentation performance for different medical image segmentation datasets \cite{khan2022leveraging}. The observation we have made is significant, emphasizing the efficiency and effectiveness of the proposed AD-Net. The finding confirms that a well-structured design can achieve outstanding results without the need for an excessively large parameter network. This includes an encoder-decoder architecture with an attention mechanism and a rational combination of loss functions. 

We introduced a method that comprises residual dilated convolutional blocks with different dilatation rates, an attention-based spatial feature enhancement block (ASFEB), and a guided decoder. Dilated convolution is a successful approach to capturing contextual information without reducing spatial resolution. Dilated convolution, by introducing gaps between kernel elements, effectively expands the receptive field, allowing the network to capture feature information in a wider scope \cite{yu2017dilated, yu2015multi, hafhouf2022improved}. This property is advantageous because skin lesions are of different sizes, shapes, and scales. where contextual information and lesion characteristics in different receptive fields are crucial for precise segmentation. In the proposed AD-Net, the dilated residual connections serve to alleviate the problem of vanishing gradients during training. 

By incorporating the ASFEB into the skip connections, the method can refine the skip connections information and enhance the lesion localization information. The guided decoder strategy facilitates the fast gradient flow and preserves important features, leading to more precise segmentation results. Through extensive experiments on diverse datasets of skin lesion images, we suggest that the proposed approach has effectiveness and improved outcomes.

In summary, our contributions include the following.
\begin{itemize}




\item  The dilated convolutional residual blocks with varying dilation rates expand the receptive field without increasing the computational burden significantly. By using varying dilation rates, the model can capture more contextual information, which is crucial for accurately delineating the boundaries, and structures of lesions.
\item  The guided decoder strategy is designed to refine the segmentation outputs progressively. It leverages intermediate features and guides the decoding process, ensuring that finer details are preserved and enhancing the overall segmentation performance.
\item  ASFEB is employed to refine feature maps within the skip connections in several ways. It effectively combines feature maps from max and average pooling layers. This combination of features helps in generating a comprehensive representation of the incoming data. By giving attention weights to the input, the model focuses more on relevant regions, such as the lesion areas, while ignoring irrelevant background information. The residual connections within ASFEB help mitigate the vanishing gradient problem and allow for more efficient training.
\item Achieving SOTA efficiency across multiple datasets like ISIC 2018, ISIC 2017, ISIC 2016, and PH2, without relying on data augmentation, represents a significant achievement. This highlights the robustness and general applicability of the proposed AD-Net.

\end{itemize}

The following sections are organised as follows:
Section 2 presents a comprehensive review of related work, highlighting key advancements and existing methodologies.
In Section 3 the details of AD-Net are explained, including an in-depth discussion of the dilated convolutional residual network, the attention-based spatial feature enhancement block, and the guided decoder strategy.
Section 4 outlines the experimental setup, including details on the datasets used, the evaluation metrics applied, and the specifics regarding implementation. Section 5 illustrates the experimental outcomes obtained from the test set on four publicly available datasets.
Section 6 presents a comprehensive discussion of the findings, including an examination of the results across different image resolutions, an analysis of computational efficiency, statistical analysis, and a discussion of the limitations of the proposed method.
The study concludes in Section 7, providing directions for future research and summarising the predominant contributions and findings.


\section{Related work}
 
In the past, older methods relied on developing specific features that could extract discriminative patterns from the image. These features were instrumental in separating the skin lesion and surrounding tissues from the rest of the image. A common approach involved the use of histogram thresholding algorithms to determine a threshold value that distinguishes the skin lesion and the surrounding tissues. These techniques tried to recognise unique patterns in the image based on changes in intensity \cite{garcia2019segmentation}.
Recent developments in neural networks have shown that they are quite effective in diagnosing skin lesions \cite{naveed2024pca}. These techniques make use of the ability of convolutional neural networks (CNNs) to independently extract discriminative characteristics from datasets, improving the robustness and efficiency of the segmentation task \cite{song2023decoupling, alahmadi2022multiscale, altan2021enhancing}. 
The authors of \cite{maji2022attention} described a generator design that improves the feature learning process of each decoder layer by using different loss functions. Feature maps produced using this method are more precise and have a deeper semantic value. To further enhance performance, attention gates are included, which is crucial for selectively engaging pertinent information. 

By integrating various modules, numerous researchers presented various methods to improve skip connection features. The authors proposed the inclusion of the spatial enhancement module in skip connections to enhance the ability to represent spatial details for semantic segmentation. Through the integration of this module into skip connections, the network adeptly captures and exploits spatial information, leading to improved segmentation performance \cite{yuan2023lightweight}.
By including attention gates in skip connections, the Attention U-Net architecture solves the semantic ambiguity problem that arises between the encoder and decoder layers. This design enables the model to selectively emphasize specific features of the encoder, facilitating better guidance and focus during the decoding phase \cite{SCHLEMPER2019197}. 
To increase accuracy and stability in medical image segmentation with the non-linear fusion of feature maps and capture advanced temporal dependencies, BCDU-Net combines U-Net and BConvLSTM at the skip connections and dense convolutions. The limitation of BCDU-Net is the increased computational complexity and memory requirements due to the incorporation of BConvLSTM and dense convolutions, which could lead to increased resource consumption and an increased risk of overfitting \cite{azad2019bi}.
The contribution of CPF-Net is the use of the global pyramid guide module (GPG) in skip connections to incorporate higher-level semantic data and a scale-aware pyramid fusion block to automatically combine features from various scales, but performance is limited due to the heavy parameters \cite{feng2020cpfnet}.\\ 
In \cite{hafhouf2022improved} the authors introduced substantial improvements to the U-Net architecture aimed at enhancing its efficacy. These enhancements have been applied to both the encoding and decoding processes. The proposed encoding pathway integrates 10 standard convolutional layers of VGG16. Additionally, it incorporates a dilated convolutional block and a pyramid pooling technique. This combination makes better spatial resolution preservation and more reliable feature extraction possible. The authors added dilated residual blocks to the decoding pipeline to further improve the segmentation maps.\\

To improve skin lesion segmentation performance, the self-attention method was used within codec components by \cite{chen2021transattunet}. 
RA-Net introduced a unique approach to skin lesion segmentation, which uses the region-aware attention mechanism to improve the effectiveness of the proposed method \cite{naveed2024ra}. 
Incorporating a channel attention strategy into a generative adversarial network suggests an improvement in the ability of the method to concentrate on particular aspects relevant to the segmentation task \cite{8832175}.
CFF-Net \cite{qin2023dynamically} is a novel technique that leverages local and global information through a local branch encoder, incorporating a CNN branch and a multi-layer perceptron branch. SUNet-DCP \cite{song2023decoupling} presents a comprehensive approach to skin lesion analysis by investigating effective techniques for feature fusion and introducing a model compression scheme to tackle the challenge of large model sizes. The objective of this approach is to improve the performance of skin lesion analysis while minimizing computational complexity and model storage requirements. 
RMMLP \cite{ji2023rmmlp} is an MLP-based method that uses dynamic matrix decomposition and rolling tensors. The model calculates a correlation matrix as a weight to guide the segmentation process and effectively combines features from different receptive fields using rolling tensors. A two-layer decoding structure with matrix decomposition integrates multi-scale information, enhancing segmentation accuracy through adaptive segmentation, feature extraction, and fusion.

\section{Methodology}

The proposed method depicted in Figure~\ref{Framework}, provides a holistic view of the individual components that make up the model.
In the segmentation of skin lesions, the network must capture variations in position, shape, scale, and hue present within different types of skin lesions. Our proposed AD-Net integrates dilated convolutional residual blocks within both the encoder and decoder components, an attention-based spatial feature enhancement block (ASFEB), and a guided decoder strategy. The enhancement of spatial features is very important because datasets have different challenges like variations in position, shape, scale, and fuzzy boundaries. To capture these variations, we employed ASFEB which refines skip connection features through attention weights, spatial information refinement, and residual learning. Ultimately, the guided decoder strategy serves to aid in the retrieval of intricate details during the decoding phase. This approach not only enhances gradient flow but also simplifies feature learning, resulting in improved segmentation performance. 
  
\subsection{Overview of the proposed method}

The proposed AD-Net consists of four sequentially dilated convolutional residual blocks, with a $2 \times 2$ maximum pooling layer after each block.
Within the encoder path, the dilated convolutional residual blocks play a crucial role in capturing features. These blocks employ a $1 \times 1$ dilated convolution and batch normalization on the residual path, by enabling the propagation of gradients across the model, our approach effectively mitigates the issue of vanishing gradients and effectively preserves the boundary information \cite{khan2022t}. This facilitates the effective propagation of information and enables the model to capture relevant features.
The dilated convolution operation is represented by Eq. \ref{dc} \cite{yu2015multi, dixit2021dilated}. 
\begin{equation}\label{dc}
{(F_{^*l}\:k)}(p) = \sum_{s+lt= p}F(s)\:k(t), 
\end{equation}
In Eq. \ref{dc} as previously indicated, the parameter $^*l$ serves as a determinant of the dilated convolution level. Specifically, when $^*l$ is set to 1, the operation functions similar to a standard convolutional operation. However, the utilization of dilated convolution introduces a distinct advantage by enabling a larger receptive field or a global perspective preserving the resolution of the image while capturing essential details from the input. This approach proves beneficial in capturing a greater amount of contextual information regarding the objects depicted in the images.\\

\begin{figure*}[htbp]
    \centering   
    \includegraphics[width = \textwidth]{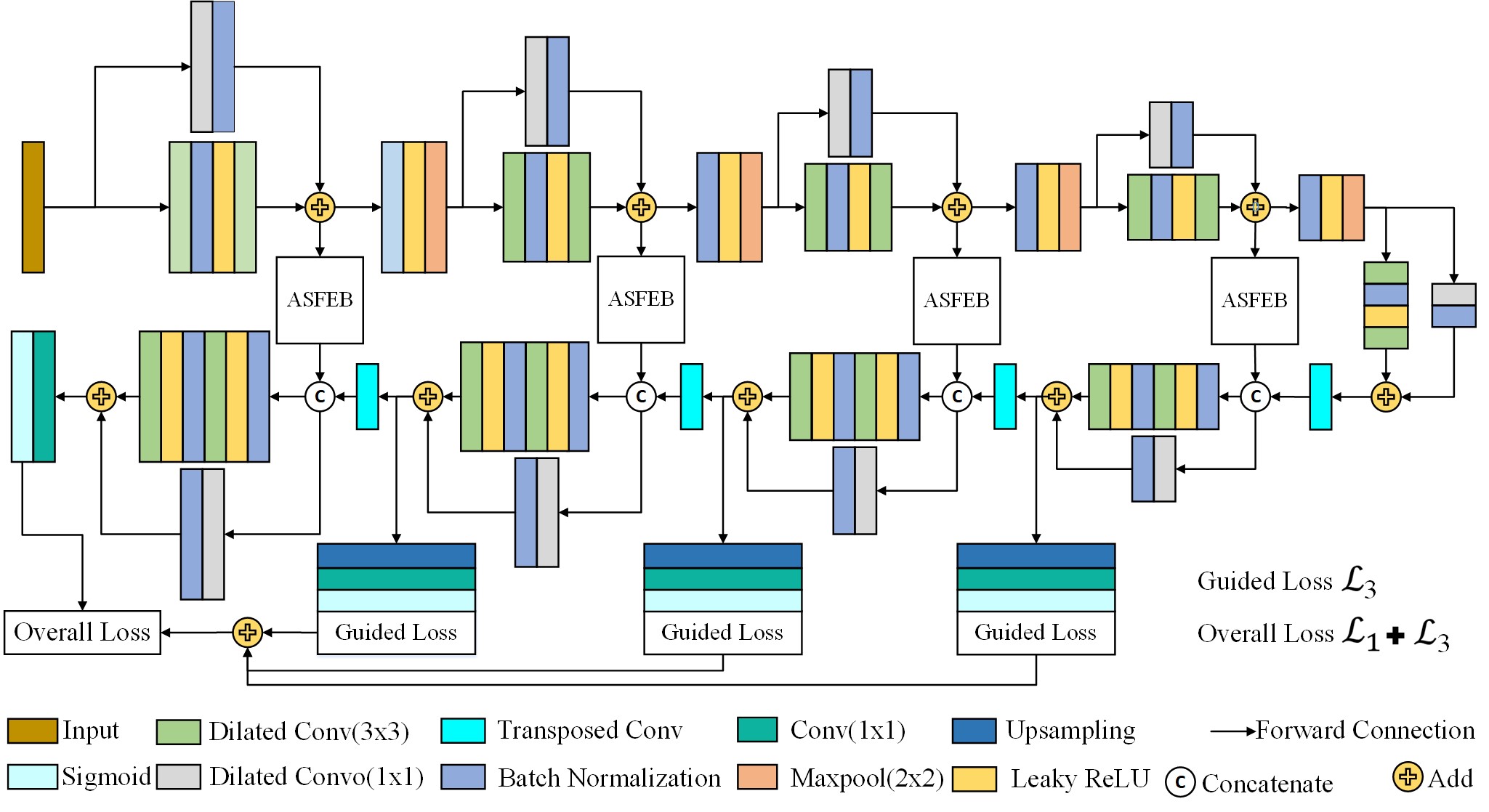}
    \caption{Overall components of the proposed AD-Net}
    \label{Framework}
\end{figure*}

The initial dilated convolutional residual block comprises 16 feature maps, each with dimensions of 256 $\times$ 256 pixels. 
Within the encoder path, as we progress through each block, the number of feature channels expands, whereas the dimensions of the feature maps undergo reduction via downsampling operations. This pattern enables the model to effectively capture features and higher-level representations as it progresses through the encoder path. Consequently, in the fourth block of the encoder path, there are 128 feature channels, each with dimensions of 32$\times$32 pixels, as shown in Figure~\ref{Framework}. In the first and second encoder blocks, the dilated rate is used $1$ for dilated convolution, which implies that the convolution operation is executed without any dilation. Consequently, the receptive field of the convolution filters remains confined to their designated size. However, the third encoder block uses a dilated convolution with a dilation rate of 2. When this dilation rate is used, the convolution filters are enabled to contain a larger receptive field, capturing information from a broader region while preserving the output size identical to the prior blocks. In this context, a dilation rate of $2$ represents a one-pixel spacing between the filter values during the application. Similarly, the fourth encoder block employs a dilation rate of $4$ for dilated convolution. This dilation rate further enhances the receptive field, enabling them to capture more spatial context information.
By increasing the dilation rate in later encoder blocks, the network can incorporate larger context information into the feature maps, enabling it to capture more global patterns and context while maintaining an effortless computational complexity.

The encoder blocks and the decoder blocks are connected via the bottleneck layer. It is also implemented by using a dilated convolutional residual block consisting of 256 feature channels, each with a size of $16 \times 16$. This layer plays a critical role in squeezing and summarising the encoded information before passing it to the decoder for further processing and reconstruction. Within the residual block of dilated convolution of the bottleneck layer, the dilated convolution layers employ a dilation rate of $4$.\\
The decoder part consists of four blocks. Each block has a residual block of dilated convolution followed by a transposed convolutional layer of kernel size 2$\times$2, which is used for upsampling purposes. The transposed convolutional layer facilitates an increase in the spatial size of the feature maps, allowing the decoder to generate high-resolution representations. This is important for recovering fine-grained details during the segmentation process. This block incorporates dilated convolutions with a dilation rate similar to the encoder path, to capture contextual information and maintain spatial resolution.

In the proposed AD-Net, a skip connection connects each decoder block to its associated encoder block. These skip connections are crucial for preserving and integrating important spatial information from the encoder to the decoder. This mechanism empowers the decoder by using both low- and high-level features, a crucial aspect for achieving precise segmentation outcomes. Additionally, the proposed AD-Net employs ASFEB at each skip connection. The purpose of ASFEB is to refine the information on skip connections, ensuring that the decoder can efficiently use the information from the corresponding encoder block for precise segmentation output. In the end, a convolutional layer 1$\times$1 is used followed by a sigmoid activation function to produce the final segmentation output. The key components of the proposed AD-Net are shown in Figure~\ref{Framework}. 

\begin{figure*}[htbp]
    \centering    
    \includegraphics[scale=0.57]{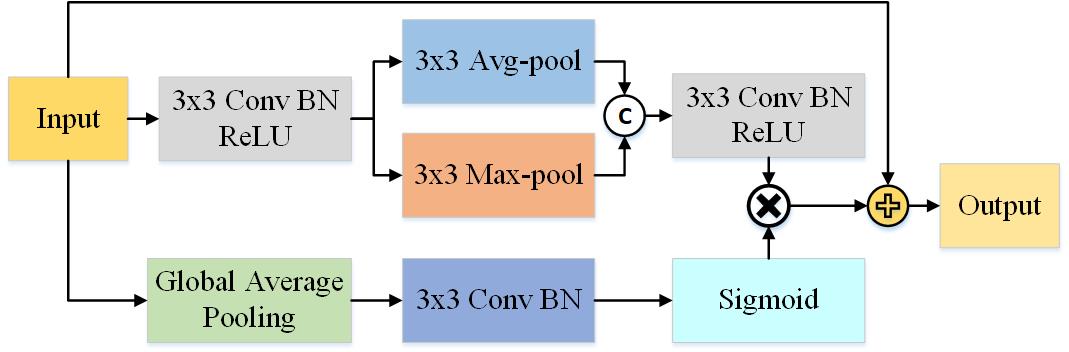}
    \caption{Attention-based Spatial Feature Enhancement Block.}
    \label{sem}
\end{figure*}

\subsection{Attention-based spatial feature enhancement block (ASFEB)}
In deep learning architecture, the pooling operations have several uses, such as reducing the size of the feature map, speeding up computations, and improving the resilience of the feature. 
For segmentation of skin lesions, capture both local details and global context is crucial, due to factors such as small size, low contrast, and the diverse colours exhibited by skin lesions. To address this, we used ASFEB in skip connections in the proposed AD-Net to enhance feature fusion, attention, spatial features information, and residual learning. 
These advantages collectively contribute to achieving more accurate segmentation outcomes, particularly in tasks, where capturing fine details and preserving object boundaries are crucial.
Figure~\ref{sem} contains an illustration of the attention-based spatial feature enhancement block.

Operations performed on the input tensor include a convolutional layer $3 \times 3$, a batch normalisation layer (BN), and a rectified linear unit layer (ReLU). Furthermore, both max pooling and average pooling operations are performed, and their results are concatenated. This process allows the method to adeptly capture both local details and global feature information, thus enhancing its ability to provide accurate predictions. The combined features then go through an additional $3 \times 3$ convolutional layer, BN, and ReLU activation. This iterative refinement of features enhances the model's ability to summarise crucial information essential for precise segmentation. In parallel, another path is introduced, which incorporates global average pooling, followed by 3 $\times$ 3 convolutional layers, BN and sigmoid activation. This path generates attention coefficients that weigh the outputs obtained from the parallel pooling operation. Ultimately, the weighted feature map is combined with the initial inputs, yielding the output. This mechanism helps to integrate the refined features with the original input, contributing to enhancing the segmentation performance. By employing both max pooling and average pooling operations in parallel, while using attention coefficients to balance their contributions, the ASFEB architecture is built to effectively catch local and global features information. 

The ASFEB technique, as described in the following equations, enhances the representation of features of the given input tensor.
\begin{equation}\label{Eq1}
{T} =  \mathbb{R}^{H\times W\times C}
\end{equation}
In Equation \ref{Eq1}, the symbol $T$ represents the given input, where $H$ denotes height, $W$ denotes width, and $C$ signifies its depth, resulting in dimensions $H \times W \times C$. These dimensions define the size and depth of the input tensor.

\begin{equation}\label{Eq2}
T_{1} = \text{ReLU}(\mu(f^{3\times3}(T)),
\end{equation}

Equation \ref{Eq2} denotes the output $T_1$, which is acquired by convolving the given input tensor $T$ with a $3 \times 3$ filter ($f^{3\times3}$), then batch normalisation ($\mu$) and $ReLU$ activation is applied.

\begin{equation}\label{Eq3}
F_{1} = (P_{m}(T_1)),
\end{equation}
\begin{equation}\label{Eq4}
F_{2} = (P_{a}(T_1)),
\end{equation}

Equations \ref{Eq3} and \ref{Eq4} represent $F_1$ and $F_2$, respectively, which are obtained by applying max pooling operation ($P_m$) and average pooling ($P_a$) operation with a stride of $3 \times 3$ on the given input $T_1$.
\begin{equation}\label{Eq5}
F_{3} = F_{1} \copyright F_{2},
\end{equation}
In Equation \ref{Eq5}, $F_3$ is derived by concatenating the max pooling ($F_1$) and average pooling ($F_2$) features.

\begin{equation}\label{Eq6}
F_{4} = \text{ReLU}(\mu(f^{3\times3}(F_{3})),
\end{equation}

Equation \ref{Eq6} shows $F_4$, obtained by applying a convolution operation ($f^{3\times3}$), subsequently batch normalisation ($\mu$) and ReLU activation ($ReLU$) on the combined features $F_3$.

\begin{equation}\label{Eq7}
F_{5} = GAP(T),
\end{equation}

In Equation \ref{Eq7}, $F_5$ is derived by performing a global average pooling operation ($GAP$) on the given input $T$.

\begin{equation}\label{Eq8}
F_{6} = \sigma(\mu(f^{3\times3} (F_{5}))),
\end{equation}

Equation \ref{Eq8} shows $F_6$, obtained by applying a convolution operation ($f^{3\times3}$), subsequently ($\mu$) and sigmoid activation ($\sigma$) in $F_5$.

\begin{equation}\label{Eq9}
F_{8} = F_{4}\otimes F_{6},
\end{equation}
$F_8$ is computed through element-wise multiplication ($\otimes$) of $F_4$ and $F_6$, as described in Equation \ref{Eq9}.

\begin{equation}\label{Eq10}
{F} = F_{8}\oplus T,
\end{equation}
Ultimately, the attention features $F$ are acquired by element-wise summation ($\oplus$) of $F_8$ and the given input $T$, as delineated in Equation \ref{Eq10}.

\subsection{Guided decoder strategy}
 
By applying individual loss functions at different layers of the decoder, the network can focus on learning specific details and features at each stage of the decoding process. This can be especially useful in tasks where fine-grained features are important for precise segmentation.
According to \cite{van2019deep}, the Jaccard loss, also referred to as the Intersection over Union (IOU) loss, measures the similarity between the predicted mask and the corresponding ground truth. This is achieved by comparing the intersection of their regions with their union. In mathematical terms, IOU loss can be expressed by equation \ref{JI}:


\begin{equation}
\label{JI}
\mathcal{L}_{Jaccard}(y,\hat{y}) = 1-\frac{\sum_{i}^{N}(y_{i} \cdot \hat{y}_{i})}{\sum_{i}^{N}(y_{i} + \hat{y}_{i} - y_{i} \cdot \hat{y}_{i}) }.
\end{equation}

Here, $y$ represents the corresponding ground truth, and $\hat{y}$ represents the predicted output. 
To enable a direct comparison with the ground truth, up-sampling is performed on the features of each decoder block to align with the input size of $256 \times 256$. This ensures consistency and facilitates an accurate evaluation of the segmentation results. This comparison helps generate improved features at these intermediate blocks, as they are trained to align with the ground truth at their respective resolutions. By incorporating the loss from each intermediate block into the final layer loss, the proposed method enhances the overall segmentation performance. This approach ensures that the model learns from multiple stages of the network and improves segmentation accuracy by considering information from different levels of feature extraction. The detailed architecture details of the proposed method, filter size, and feature maps are presented in Table~\ref{architecture details}. These values are set empirically and the results are shown in Table~\ref{tab:ablation17dilationrate}.

\begin{table*}
  \centering
  \caption{Proposed architecture layers details}
  \adjustbox{scale = 0.8}{
    \begin{tabular}{cccc}
    \hline
    \multicolumn{1}{|l|}{Block name} & \multicolumn{1}{l|}{Layers details } & \multicolumn{1}{l|}{Filter size} & \multicolumn{1}{l|}{Feature maps} \\
    \hline
    \multicolumn{1}{l}{Input Layer} & $256\times256\times3$ &       &  \\
    \hline
    \multicolumn{1}{l}{Encoder Block-1} & conv 2d \textsuperscript{1} & \multicolumn{1}{l}{3x3} & 16 \\
          & BN \textsuperscript{2}&       &  \\
          & conv 2d & \multicolumn{1}{l}{3x3} & 16 \\
          & leakyReLU+BN &       &  \\
          & conv 2d at residual path& \multicolumn{1}{l}{1x1} & 16 \\
          & BN+leakyReLU &       &  \\
    &{Max Pooling} \\
    \hline
   \multicolumn{1}{l}{Encoder Block-2} & conv 2d & \multicolumn{1}{l}{3x3} & 32 \\
          & BN &       &  \\
          & conv 2d & \multicolumn{1}{l}{3x3} & 32 \\
          & leakyReLU+BN &       &  \\
          & conv 2d at residual path& \multicolumn{1}{l}{1x1} & 32 \\
          & BN+leakyReLU &       &  \\
    &{Max Pooling} \\
    \hline
    \multicolumn{1}{l}{Encoder Block-3} & conv 2d & \multicolumn{1}{l}{3x3} & 64 \\
          & BN &       &  \\
          & conv 2d & \multicolumn{1}{l}{3x3} & 64 \\
          & leakyReLU+BN &       &  \\
          &conv 2d at residual path& \multicolumn{1}{l}{1x1} & 64 \\
          & BN+leakyReLU &       &  \\
    &{Max Pooling} \\
    \hline
    \multicolumn{1}{l}{Encoder Block-4} & conv 2d & \multicolumn{1}{l}{3x3} & 128 \\
          & BN &       &  \\
          & conv 2d & \multicolumn{1}{l}{3x3} & 128 \\
          & leakyReLU+BN &       &  \\
          & conv 2d at residual path& \multicolumn{1}{l}{1x1} & 128 \\
          & BN+leakyReLU &       &  \\
    &{Max Pooling} \\
    \hline
    \multicolumn{1}{l}{Bottleneck} & conv 2d & \multicolumn{1}{l}{3x3} & 256 \\
          & BN+ LeakyReLU &       &  \\
          & conv 2d at residual path& \multicolumn{1}{l}{1x1} & 256 \\
          
    \hline
    \end{tabular}%
    }\\
\noindent{\footnotesize{ \textsuperscript{1} conv 2d: dilated convolution}}\\
\noindent{\footnotesize{ \textsuperscript{2}  BN: Batch Normalization}}
  \label{architecture details}%
\end{table*}%

\subsection{Loss functions}
The segmentation of skin lesions poses challenges due to several intricate factors, for example, fuzzy boundaries, low contrast, irregular shapes, and interference of various elements such as ink marks, blood vessels, ruler imprints, and hair. In this study, two training strategies were employed: (1) using a single loss function and (2) combining two loss functions. 
Through the combination of various loss functions, the training process gains advantages in terms of the class imbalance problem, improves localization accuracy, and mitigates false positives and false negatives, thereby enhancing overall performance.
The proposed AD-Net undergoes training using a combination of loss functions: binary cross-entropy (BCE) loss \cite{jadon2020survey}, Dice coefficient (Dice) loss \cite{jadon2020survey}, and focal Tversky (FTL) loss \cite{jadon2020survey}.
Using this set of loss functions during training, the approach takes advantage of the unique strengths and characteristics of each loss function. This combination can potentially improve both the performance and the optimization of the model.

The mathematical formulations for BCE loss, Jaccard loss, Dice loss and Focal Tversky loss are represented in Eq.~\ref{bce}, Eq.~\ref{JI}, Eq.~\ref{diceloss}, and Eq.~\ref{ftl}, respectively.
The relationship between each estimated probability and the corresponding class output is evaluated using the BCE loss \cite{jadon2020survey}.
\begin{equation}
\label{bce}
\mathcal{L}_{BCE}(y,\hat{y}) = -\sum_{i=1}^{N} ({y}_{i}\mathrm{log}{\hat{y}}_{i}  + (1-{y}_{i})\mathrm{log}(1-{\hat{y}}_{i})).
\end{equation}

\begin{equation}
\label{diceloss}
\mathcal{L}_{DICE} (y,\hat{y}) = 1-\frac{2\sum_{i}^{N}(y_{i} \cdot \hat{y}_{i})}{\sum_{i}^{N}y^2_{i}+ \sum_{i}^{N}\hat{y}^2_{i}},
\end{equation}

\begin{equation}
TI_{c} = \frac{\sum_{i}^{N}(y_{i} \cdot \hat{y}_{i})}{\sum_{i}^{N}(y_{i} \cdot \hat{y}_{i}) +\alpha\sum_{i}^{N}(y_{i}\cdot(1-\hat{y}_{i})) +\beta\sum_{i}^{N}((1-y_{i})\cdot\hat{y}_{i})},
\end{equation}

Where $y$ denotes the ground truth, $\hat{y}$ represents the predicted output, $N$ is the count of samples, $\alpha$ and $\beta$ are hyperparameters.
To convert the Tversky index into a loss function, the complement of the Tversky index can be minimized. The FTL is defined as:

\begin{equation}
\label{ftl}
\mathcal{L}_{FTL}  = \sum_{c}^{} (1-TI_{c})^{\frac{1}{\gamma}},
\end{equation}
where $\gamma$ is a hyper-parameter that can be adjusted in the range [1-3]. By incorporating the focal mechanism, the FTL is designed to assign more weight to false negatives and false positives, which can significantly improve a model's performance, particularly on imbalanced datasets.\\

The combined loss function for option 1 is defined as:
\begin{equation}
\mathcal{L}\textrm{1} = \mathcal{L}_{BCE} + \mathcal{L}_{FTL},
\end{equation}
For Option 2, the loss function is explained as:
\begin{equation}
\mathcal{L}\textrm{2} = \mathcal{L}_{BCE} + \mathcal{L}_{DICE},
\end{equation}
For the guided decoder, we employed the Jaccard loss after each decoder block. The loss function is expressed as:

\begin{equation}
\mathcal{L}\textrm{3} = \mathcal{L}_{Jaccard},
\end{equation}

The overall loss functions (LF) to train the model are defined in Eq.~\ref{totalloss} and Eq.~\ref{totallossA}.

\begin{equation}\label{totalloss}
\mathcal{L}_{\text{A}} =  \mathcal{L}\textrm{1}+ \mathcal{L}\textrm{3},
\end{equation}

\begin{equation}\label{totallossA}
\mathcal{L}_{\text{B}} =  \mathcal{L}\textrm{2}+ \mathcal{L}\textrm{3},
\end{equation}
\begin{table*}[htbp]
  \centering
  \caption{Results of different combinations of loss functions on the baseline method for the ISIC 2017 dataset}
    \adjustbox{max width=\textwidth}{
    \begin{tabular}{lccccccc}
    \hline
     \textbf{Model name}& \textbf{Parameters} & \textbf{LF}  & {\textbf{J}} & {\textbf{D}} & {\textbf{Acc}} & {\textbf{Sn}} & {\textbf{Sp}} \\
    \hline   
    baseline & 0.488M & $\mathcal{L}_{DICE}$ & 72.12 & 81.27 & 92.00 & 79.33 & 97.44 \\
    baseline & 0.488M & $\mathcal{L}_{FTL}$ & 73.26 & 82.50 & 92.44 & 84.35 & 95.64 \\
    baseline & 0.488M & $\mathcal{L}\textrm{2}$ & 82.20 & 89.28 & 95.03 & 89.02 & 95.20 \\
    baseline &0.488M & $\mathcal{L}\textrm{1}$ & 82.26 & 89.32 & 95.09 & 89.14 & 95.21 \\  
    \hline
    \end{tabular}%
    }
  \label{tab:lossresults}%
\end{table*}
\begin{figure*}[htbp]
    \centering
    \includegraphics[width=\textwidth]{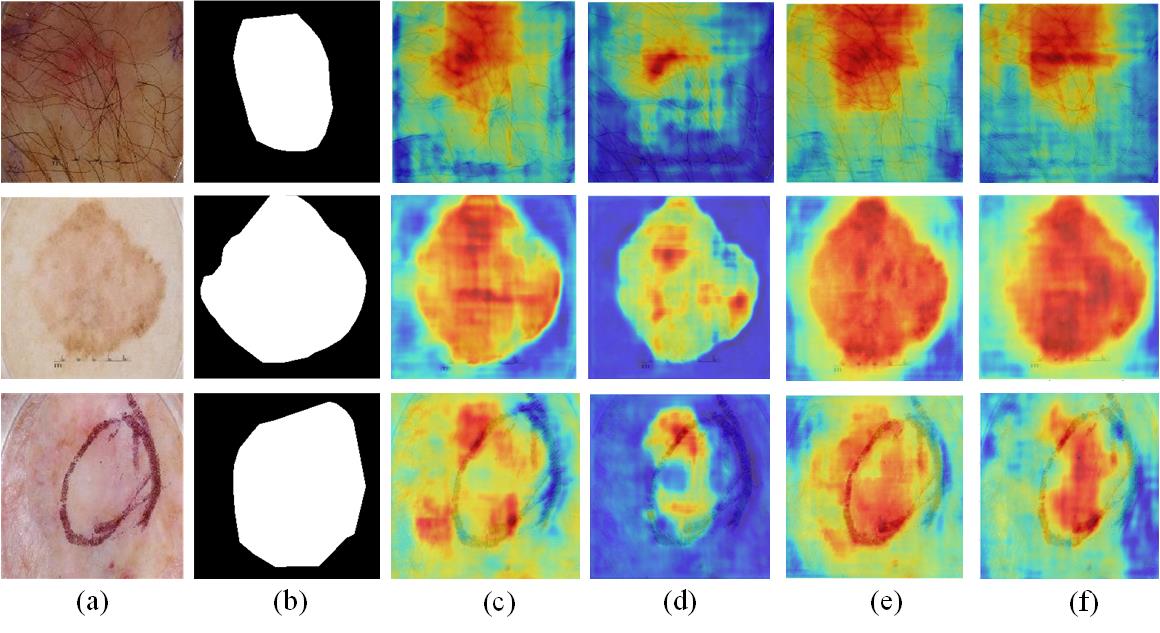}
    \caption{Heat maps visualization of the baseline method for different loss functions: (a) skin lesion images, (b) ground truth masks, (c) Heat maps of FTL loss, (d) Heat maps of dice loss, (e) Heat maps of $\mathcal{L}\textrm{1}$ loss, and (f) Heat maps of $\mathcal{L}\textrm{2}$ loss}
    \label{visualloss}
\end{figure*}

The impact of individual and combined loss functions on the baseline model is evaluated, with the outcomes presented in Table~\ref{tab:lossresults} and Figure~\ref{visualloss}. The Jaccard (J), Dice (D), Accuracy (Acc), Sensitivity (Sn), and Specificity (Sp) are used to measure the performance of the proposed model. These outcomes deliver useful insights into the performance and effectiveness of different loss functions for skin lesion segmentation. Figure~\ref{visualloss} includes heat maps generated using gradient-weighted class activation mapping (Grad-CAM) \cite{selvaraju2017grad} on the ISIC 2017 dataset. This Figure~\ref{visualloss} illustrates the model's learning outcomes and its ability to segment lesions under various loss function configurations. The heat maps demonstrate that the combined loss functions strategy leads to more precise and consistent identification of lesion area, highlighting the effectiveness of this approach in skin lesion segmentation. \\

\begin{figure*}[htbp]
    \centering
    \includegraphics[width=\textwidth]{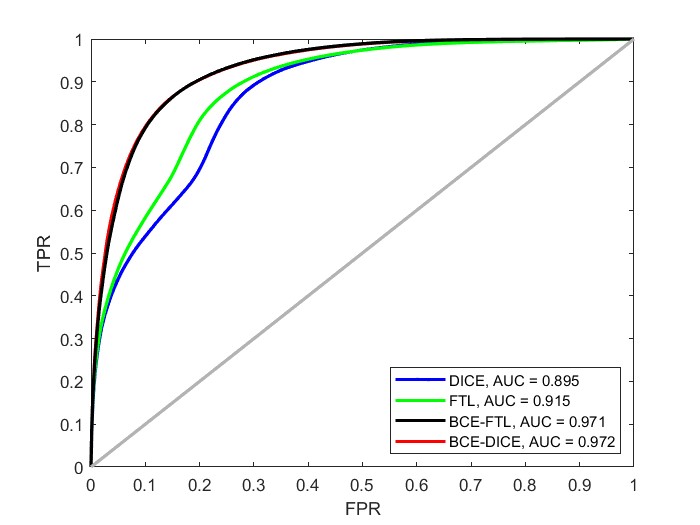}
    \caption{Comparison of the ROC curves for different loss functions on baseline method}
     \label{ROC}
\end{figure*}

The performance of the AD-Net was also evaluated using the receiver operating characteristic (ROC) curve. This curve illustrates the relationship between the true positive rate (sensitivity) and the false positive rate (1 - specificity) across different threshold settings. The primary objective of this study is to segment the lesion region, designating the lesion area with a value of 1 and the non-lesion area with a value of 0. The ROC curve is widely regarded as the most effective tool for assessing the separability of classes. Each point on the curve represents the performance of the classifier at a specific threshold. The proposed AD-Net ability to distinguish between classes is quantified by the area under the ROC curve (AUC). A higher AUC indicates a greater accuracy in class separation. In our case, the proposed method achieved an AUC of 0.972, demonstrating its exceptional capability to differentiate between the two classes (lesion and background) \cite{altan2022deepoct, zafar2020skin}. Figure~\ref{ROC} displays ROC curves of the proposed method when employing single and combined loss functions on the ISIC 2017 dataset.


\section{Experiments}
This section presents the details of the employed dataset, the performance evaluation metrics utilized, the implementation details, and the ablation experiments.

\subsection{Details of datasets}
The performance of AD-Net is evaluated on public benchmark datasets. These datasets are ISIC 2016 \cite{gutman2016skin}, ISIC 2017 \cite{codella2018skin}, ISIC 2018 \cite{tschandl2018ham10000,codella2019skin}, and PH2 \cite{mendoncca2013ph}.
These datasets are widely recognized and commonly utilized for skin cancer analysis. They offer diverse examples of skin lesion images, enabling comprehensive evaluation and comparison of different methods.
ISIC 2018 dataset consists of dermoscopic images, sourced from various clinical centers. It comprises 2594 images for training and an additional 1000 images for assessing model performance during testing. Similarly, the ISIC 2017 dataset consists of 2000 images, the validation set consists of 150, and the test set consists of 600 images. The ISIC 2016 dataset consists of 900 skin lesion images for training and 379 skin lesion images for testing. 
PH2 is a collection of dermoscopic images primarily aimed at melanoma detection and diagnosis. It includes 200 skin lesion images with corresponding ground truth masks.
The particulars of these datasets are shown in Table~\ref{datasets}.
\begin{table*}[htbp]
  \centering
  \caption{Details of the datasets for skin lesions segmentation}
   \adjustbox{max width=\textwidth}{
    \begin{tabular}{ccccc}
    \hline
    \multirow{2}[4]{*}{\textbf{Datasets}} & \multicolumn{3}{c}{\textbf{Number of Images}} & 
    \multirow{2}[4]{*}{\textbf{Image Resolution Range}} \\
\cmidrule{2-4}          & \textbf{Training} & \textbf{Validation} & \textbf{Test} &  \\
    \hline
    {ISIC 2016} & {900} & N.A & {379} & {679$\times$566 - 2848$\times$4288} \\
   
    {ISIC 2017} & {2000} & {150} & {600} & {679$\times$453 - 6748$\times$4499} \\
    
    {ISIC 2018} & {2594} & N.A & 1000 & {679$\times$453 - 6748$\times$4499} \\
    {PH2} & {200} & N.A & N.A & {768$\times$560} \\
    \hline
    \end{tabular}%
    }
  \label{datasets}%
\end{table*}%

\subsection{Evaluation criteria}

As described in \cite{gutman2016skin}, the performance evaluation of the proposed technique uses five key assessment metrics: accuracy, sensitivity, Jaccard index, dice coefficient, and specificity.
The skin lesion segmentation evaluation criteria were selected based on the guidance provided by the ISIC competition leaderboard, a well-known platform. Using these assessment measures, a comprehensive analysis of the performance of the model can be conducted. These measures provide valuable information on various aspects of the segmentation results. The choice to utilize these specific criteria, with a particular emphasis on the Jaccard index (IOU) as the primary metric, aligns with the guidance provided by the ISIC challenge leaderboard, as in \cite{codella2019skin}. 

\begin{equation}\label{}
\mathrm{Accuracy} = \frac{{T_P+T_N}}{{T_P+T_N+F_P+F_N}}
\end{equation}
\begin{equation}\label{}
\mathrm{Sensitivity} = \frac{{T_P}}{{T_P + F_N}}
\end{equation}
\begin{equation}\label{}
\mathrm{IOU} = \frac{{T_P}}{{T_P+ F_P +F_N}}
\end{equation}
\begin{equation}\label{}
\mathrm{Dice/F1} = \frac{{2*T_P}}{{2*T_P+ F_P +F_N}}
\end{equation}
\begin{equation}\label{}
\mathrm{Specificity} = \frac{{T_N}}{{T_N + F_P}}
\end{equation}

\subsection{\textbf{Implementation details}}

The implementation specifics of the proposed AD-Net are detailed in this section, utilizing several widely used benchmark datasets, including PH2, ISIC 2016, ISIC 2017, and ISIC 2018. These datasets are widely recognized as standard benchmarks in the field, enabling the evaluation of our method's performance and its ability to generalize across diverse datasets.
In our experimental configuration, we standardized the dimensions of all datasets to $256\times 256$ pixels. To train AD-Net, a total of twenty percent of the training data was set aside for validation. We utilized the Adam optimizer \cite{kingma2014adam} for training, employing mixed loss functions to enhance the model performance. If the validation set performance does not improve, the Adam optimizer's learning rate decreases by a factor of 0.25 after four epochs, starting at 0.001. We also used the early stop strategy to handle the overfitting issue and to calculate the maximum number of training epochs dynamically. We employed a batch size of 10 for the ISIC 2016 dataset, 8 for the ISIC 2017 dataset, and 8 for the ISIC 2018 dataset. However, AD-Net achieved SOTA performance without requiring additional data.\\
The AD-Net is implemented in Keras and TensorFlow as the back end. All variants of the model are trained on the NVIDIA K80 GPU with the following specifications: An Intel Xeon CPU running at 2.20 GHz, 13 GB of RAM, and a Tesla K80 accelerator with 12 GB of GDDR5 VRAM make up the GPU runtime environment.

\begin{table*}[htbp]
  \centering
  \caption{Results of the ablation study showing the impact of various components on the ISIC 2017 dataset}
    \adjustbox{max width=\textwidth}{
    \begin{tabular}{lccccccc}
    \hline
    \textbf{Model name}& \textbf{LF} & \textbf{Parameters}  & {\textbf{J}} & {\textbf{D}} & {\textbf{Acc}} & {\textbf{Sn}} & {\textbf{Sp}} \\
    \hline  
    baseline & $\mathcal{L}\textrm{2}$ & 1.95 & 81.63 & 88.81 & 94.89 & 88.56 & 95.75 \\
    baseline & $\mathcal{L}\textrm{1}$ & 1.95 & 82.27 & 89.38 & 95.04 & 89.00 & 95.58 \\
    baseline + ASFEB at skip connections & $\mathcal{L}\textrm{2}$ & 2.82 & 83.44 & 90.07 & 95.41 & 89.32 & 96.73 \\
    baseline + ASFEB at skip connections &  $\mathcal{L}\textrm{1}$ & 2.82 & 84.22 & 90.66 & 95.61 & 90.14 & 96.56\ \\
    baseline + ASFEB + guided loss & $\mathcal{L}_\textrm{B}$ & 2.92 & 84.15 & 90.53 & 95.53 & 89.27 & \textbf{97.36} \\
     
    baseline + ASFEB + guided loss  &  $\mathcal{L}_\textrm{A}$ & 2.92 & \textbf{84.51} & \textbf{90.86} & \textbf{95.82} & \textbf{90.22} & 96.47 \\
    \hline
    \end{tabular}%
    }
    \label{tab:ablation17}%
\end{table*}%

\begin{figure*}[htbp]
    \centering
    \includegraphics[width = \textwidth]{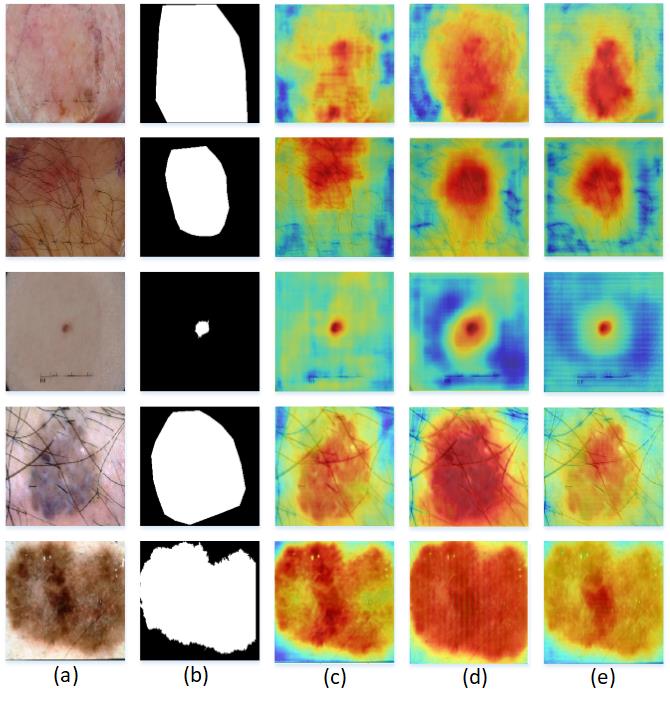}
    \caption{Heat maps of AD-Net for different components: (a) input images, (b) ground truths, (c) heat maps of baseline method, (d) heat maps of baseline method with ASFEB, (e) heat maps of baseline method, ASFEB, and guided loss strategy}
    \label{attentionmapsvisual}
\end{figure*}

\begin{table*}[htbp]
  \centering
  \caption{Outcomes of ablation study on ISIC 2017 dataset with different numbers of trainable parameters}
    \adjustbox{max width=\textwidth}{
    \begin{tabular}{lccccccc}
    \hline
    \textbf{Model name}& \textbf{LF} & \textbf{Parameters}  & {\textbf{J}} & {\textbf{D}} & {\textbf{Acc}} & {\textbf{Sn}} & {\textbf{Sp}} \\
    
    \hline  
    
    baseline + ASFEB + guided loss & $\mathcal{L}_\textrm{A}$ & 0.71 & 83.54 & 90.08 & 95.33 & 89.34 & 96.92 \\
    baseline + ASFEB + guided loss &  $\mathcal{L}_\textrm{A}$ & 2.92 & \textbf{84.51} & \textbf{90.86} & \textbf{95.82} & \textbf{90.22} & 96.47 \\
    baseline + ASFEB + guided loss &  $\mathcal{L}_\textrm{A}$ & 11.22 & 82.83 & 89.39 & 94.97 & 88.47 & 97.34\ \\
    baseline + ASFEB + guided loss  &  $\mathcal{L}_\textrm{A}$ & 44.99 & 83.91 & {90.29} & 95.25 & 89.22 & \textbf{97.54} \\
    \hline
    \end{tabular}%
    }
  \label{tab:ablation17parameters}%
\end{table*}%

\subsection{\textbf{Ablation study on the ISIC 2017 dataset}}
We conducted an ablation investigation utilizing various components to evaluate AD-Net's capabilities. Various experiments were performed, including those with single and combined loss functions, using the baseline method. The trainable parameters in millions were also taken into account during the analysis. These experiments aimed to evaluate the contribution and significance of each component in achieving SOTA results.
We introduced ASFEB at skip connections and a guided decoder strategy in the proposed method for improved performance. Table~\ref{tab:ablation17} shows the results of the ablation study for the ISIC 2017 dataset for each component. Furthermore, Figure~\ref{attentionmapsvisual} illustrates the heat maps for all elements of AD-Net. 
Observing performance, we used the loss $\mathcal{L}_{\text{A}}$ to train the model on all data sets.\\ 

More ablation investigations were conducted to determine the appropriate parameter values for the segmentation of the skin lesions. Table~\ref{tab:ablation17parameters} presents the results, which provide insight into the performance of the proposed method in different combinations of parameters.
 Furthermore, Figure~\ref{parameterspsa} visually shows the connection between the Jaccard index and the count of trainable parameters, highlighting the influence of parameter choices on segmentation performance. These insights are valuable in improving and fine-tuning the proposed method to achieve precise and efficient skin lesion segmentation outcomes.
Furthermore, a comparison of two optimizers was conducted. The outcomes are displayed in Table~\ref{tab:ablation17optimizer}, indicating that the Adam optimizer performs better compared to stochastic gradient descent (SGD). This implies that the Adam optimizer is better suited for optimizing AD-Net and can enhance segmentation performance.
\begin{table*}[htbp]
  \centering
  \caption{Outcomes of AD-Net on ISIC 2017 dataset with SGD and Adam optimizers. Best performance denoted by bold numbers in the table}
    \adjustbox{max width=\textwidth}{
    \begin{tabular}{lccccccc}
    \hline
    \textbf{Model name}& \textbf{LF} & \textbf{Optimizer}  & {\textbf{J}} & {\textbf{D}} & {\textbf{Acc}} & {\textbf{Sn}} & {\textbf{Sp}} \\    \hline  
    
    Proposed method & $\mathcal{L}_\textrm{A}$ & SGD & 83.71 & 90.23 & 95.52 & 89.83 & \textbf{96.58} \\
    Proposed method & $\mathcal{L}_\textrm{A}$&  Adam & \textbf{84.51} & \textbf{90.86} & \textbf{95.82} & \textbf{90.22} & 96.47 \\
 
    \hline
    \end{tabular}%
    }
  \label{tab:ablation17optimizer}%
\end{table*}%

Table~\ref{tab:ablation17dilationrate} presents a comparison of the proposed AD-Net with various dilation convolution filter sizes. The results indicate that with 2.92 million trainable parameters, the 3$\times$3 filter size yields the best performance. The choice of filter size in dilated convolutions significantly influences the receptive field, the resolution of the features, the computational efficiency, and the overall performance of the proposed method \cite{yu2017dilated, yu2015multi, hafhouf2022improved}. By conducting empirical evaluations, we choose the $3\times3$ filter size for AD-Net.

\begin{table*}[htbp]
  \centering
  \caption{{Results of the AD-Net using different filter sizes on the ISIC 2017 dataset}}
    \adjustbox{max width=\textwidth}{
    \begin{tabular}{lcccccccc}
    \hline
    \textbf{Model name}& \textbf{Filter size} & \textbf{LF} & \textbf{Parameters}  & {\textbf{J}} & {\textbf{D}} & {\textbf{Acc}} & {\textbf{Sn}} & {\textbf{Sp}} \\
    \hline  
    
    Proposed method & $1\times1$ &$\mathcal{L}_\textrm{A}$ & 1.35 & 75.94 & 84.21 & 93.18 & 84.43 & 94.55 \\
    Proposed method & $3\times3$ &$\mathcal{L}_\textrm{A}$ & 2.92 & \textbf{84.51} & \textbf{90.86} & \textbf{95.82} & {90.22} & 96.47 \\
    Proposed method & $5\times5$& $\mathcal{L}_\textrm{A}$ & 6.1 & 84.40 & 90.71 & 95.76 & \textbf{90.68}  & 95.96\ \\
  
    \hline
    \end{tabular}%
    }
  \label{tab:ablation17dilationrate}%
\end{table*}%

\begin{table*}[htbp]
  \centering
   
    \caption {Comparing the proposed AD-Net's performance (mean $\pm$ standard deviation) against alternative SOTA techniques using the ISIC 2018 dataset. Numbers in bold indicate the best performance.}
  \adjustbox{max width=\textwidth}{
    \begin{tabular}{lccccccc}
    \hline
    \textbf{Model name} & \textbf{Parameters}  & {\textbf{J}} & {\textbf{D}} & {\textbf{Acc}} & {\textbf{Sn}} & {\textbf{Sp}} \\
    \hline
    LeaNet \cite{hu2024leanet} & 0.11  & 78.39 & 88.25 & 94.72 & 91.03 & 98.24 \\
    CPFNet \cite{9049412} & 43.30 & 79.88 & 87.69 & 94.96 & 89.53 & 96.55 \\ 
    DAGAN \cite{LEI2020101716}  & 56  &81.13 & 88.07 & 93.24 & 90.72 & 95.88 \\
    FAT-Net \cite{WU2022102327}  & 30  &82.02 & 89.03 & 95.78 & 91.00 & \textbf{96.99} \\
    CFF-Net \cite{qin2023dynamically}  & 9.71 & 82.55 & 90.08  & --- & 88.63 & ----\\
    AS-Net \cite{HU2022117112}   & 24.90  &83.09 &  89.55 & 95.68 & 93.06 & 94.69 \\ 
    SLSN \cite{dong2023learning} &3.3& 83.73  & 90.54 & \textbf{96.47} & 91.00 & --- \\
    {ADF-Net} \cite{huang2024adf} & 36.21 & 84.96 & 91.12 & 96.83 & 92.68 & 97.67 \\
     RMMLP \cite{ji2023rmmlp}  &13.91  &85.40  & 91.98 & --- & --- &  ---\\
    Ms RED \cite{DAI2022102293} & 3.80  &83.86  & 90.33 & 96.45 & 91.10 & --- \\
    \hdashline
    U-Net \cite{ronneberger2015u} & 31.10  & 83.66 $\pm$ {0.152}  & 90.16 $\pm$ {0.116} & 94.00 $\pm$ {.083} & 90.93 $\pm$ {0.094} & 91.81 $\pm$ {0.168} \\
    UNet++ \cite{zhou2018unet++} & 9.20  & 84.83 $\pm$ {0.154} & 90.86 $\pm$ {0.114} & 94.38 $\pm$ {0.090} & 91.72 $\pm$ {0.095} & 93.17 $\pm$ {0.158}\\
    Swin-Unet \cite{cao2023swin}  & 2.07 & 85.31 $\pm$ {0.130} &  91.39 $\pm$ {0.099} & 94.77 $\pm$ {0.080} & 91.89 $\pm$ {0.086} & 93.31 $\pm$ {0.148} \\
    ARU-GD \cite{maji2022attention}  & 33.50  & 85.97 $\pm$ {0.134} & 91.78 $\pm$ {0.095} & 95.01 $\pm$ {0.078} & 92.82 $\pm$ {0.089} & 94.13 $\pm$ {0.136} \\ 
    \hline
     {Proposed} & \textbf{2.92}  &\textbf{87.39 } $\pm$ {0.139} & \textbf{92.53} $\pm $ {0.102} & 95.64 $\pm$ {0.074} & \textbf{93.15} $\pm $ {0.087} & 94.92 $\pm$ {0.132} \\
   
    \hline
    \end{tabular}%
    }
  \label{tab:addlabel18}%
\end{table*}%

\begin{figure*}[htbp]
    \centering
    \includegraphics[width=\textwidth]{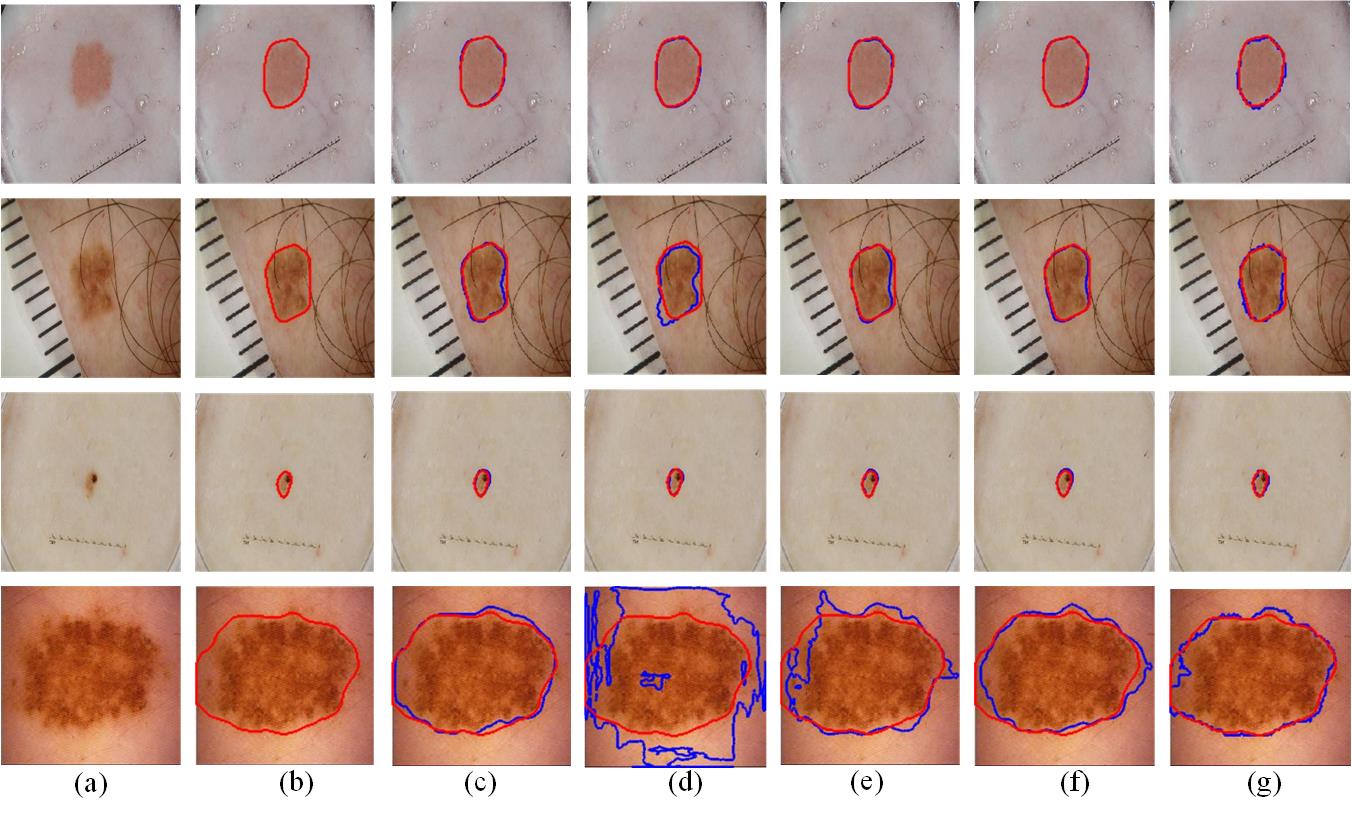}
  \caption{ISIC 2018 dataset visual results: (a) skin lesion images (b) ground truth masks (c) Proposed AD-Net (d) U-Net (e) U-Net++ (f) ARU-GD (g) Swin-Unet. The red contours show the ground truths and the blue contours show the segmentation outcomes}
    \label{Fig: ISIC2018attentionmaps}
\end{figure*}
\section{Results}
This section details the results of the AD-Net across four datasets, providing insights into computational complexity and presenting a thorough statistical analysis.

\subsection{\textbf{ISIC 2018 dataset comparison with benchmark models}}

Based on the evaluation using the ISIC 2018 dataset, our AD-Net was compared with 14 SOTA methods, including LeaNet, CPFNet, DAGAN, FAT-Net, CFF-Net, AS-Net, U-Net \footnote{ Code available at \url{ https://github.com/zhixuhao/unet}}, SLSN, Ms RED, Unet++ \footnote{ Code available at \url{https://github.com/MrGiovanni/UNetPlusPlus}}, ADF Net, Swin-Unet \footnote{Code available at \url{https://github.com/HuCaoFighting/Swin-Unet}}, RMMLP and ARU-GD \footnote{ Code available at \url{https://github.com/dhirajmaji7/Attention-Res-Unet}}. The results presented in Table~\ref{tab:addlabel18} show that our method achieved improvements of 11.48\%, 9.40\%, 7.72\%, 6.55\%, 5.86\%, 5.18\%, 4.46\%, 4.37\%, 4.21\%, 3.02\%, 2.86\%, 2.44\%, 2.33\%, and 1.65\% in terms of the Jaccard index compared to these SOTA methods.

In addition, a detailed comparison was performed with other widely recognised methods, namely U-Net, Unet++, Swin-Unet and ARU-GD, training them, and the results are summarised in Table~\ref{tab:addlabel18}. Each assessment metric shown in Table~\ref{tab:addlabel18} consists of the mean and standard deviation of the test data set for the U-Net, Unet ++, Swin-Unet and ARU-GD models. The other methods in Table~\ref{tab:addlabel18} are the articles cited in the skin lesion segmentation domain. These comparisons provide a comprehensive analysis of the efficiency and performance gains achieved by our AD-Net.

In addition, the visual results show different challenges in skin lesion segmentation, such as irregular shapes, low contrast, the presence of artefacts, and small lesions. Figure~\ref{Fig: ISIC2018attentionmaps} presents these visual results, highlighting the robustness of AD-Net in handling various sizes and irregular shapes of skin lesions, thus demonstrating state-of-the-art performance on unseen test data.

\begin{table*}[htbp]
  \centering
  \caption{{Comparing the proposed AD-Net's performance (mean $\pm$ standard deviation) against alternative SOTA techniques using the ISIC 2017 dataset. Numbers in bold indicate the best performance.}}
  \adjustbox{max width=\textwidth}{
    \begin{tabular}{lcccccc}
    \hline
    \textbf{Model name} & \textbf{Parameters}  & {\textbf{J}} & {\textbf{D}} & {\textbf{Acc}} & {\textbf{Sn}} & {\textbf{Sp}} \\
    \hline
    DAGAN \cite{LEI2020101716} & 56 & 75.94 & 84.25 & 93.26 & 83.63 & 97.25 \\
    SUNet-DCP \cite{song2023decoupling}  & 24.6 & 76.00 & 84.60  & 95.60 & 84.90 & 98.05 \\
    FAT-Net \cite{WU2022102327} & 30 & 76.53 & 85.00  & 93.26 & 83.92 & \textbf{97.25} \\
    {TMAHU-Net} \cite{dong2024transformer} & 37.43 & 77.11 & 87.15  & --- & \textbf{91.22} & --- \\
    RMMLP \cite{ji2023rmmlp}  &13.91  & 78.33  & 86.60 & --- & --- &  ---\\
    Ms RED \cite{DAI2022102293} & 3.80 & 78.55 & 86.48 & 94.10 & ---  & --- \\
    {ADF-Net} \cite{huang2024adf} & 36.21 & 78.92 & 86.79 & 94.52 & 85.49 & 97.68 \\
    {LeaNet} \cite{hu2024leanet} & 0.11  & 78.93 & 88.89 & 95.72 & 90.63 & 97.72 \\
    AS-Net \cite{HU2022117112} & 24.90 & 80.51 & 88.07 & 94.66 & 89.92 & 95.72 \\
    CFF-Net \cite{qin2023dynamically}  & 9.71 & 81.07 & 89.09  & --- & 86.56 & ----\\
    \hdashline
    U-Net \cite{ronneberger2015u} & 31.10 & 75.69 $\pm$ {0.208} & 84.12 $\pm$ {0.175} & 93.29 $\pm$ {0.089} & 84.30 $\pm ${0.143} & 93.41 $\pm$ {0.126} \\
    UNet++ \cite{zhou2018unet++} & 9.20 & 78.58 $\pm$ {0.191} & 86.35 $\pm$ {0.159} & 93.73 $\pm$ {0.087} & 87.13 $\pm$ {0.119} & 94.41 $\pm$ {0.107}\\
    ARU-GD \cite{maji2022attention} & 33.50 & 80.77 $\pm$ {0.159} & 87.89 $\pm$ {0.126} & 93.88 $\pm$ {0.078} & 88.31 $\pm$ {0.114} & 96.31 $\pm$ {0.085}\\
    Swin-Unet \cite{cao2023swin} & 2.07  & 80.79 $\pm$ {0.158} &  88.27 $\pm$ {0.129} & 94.53 $\pm$ {0.079} & 89.35 $\pm$ {0.094} & 94.96 $\pm$ {0.099} \\
    \hline
    {Proposed } & \textbf{2.92} & \textbf{84.51} $\pm$ {0.135} & \textbf{90.86} $\pm $ {0.104} & \textbf{95.82} $\pm $ {0.068} & 90.22 $\pm $ {0.097} & 96.47 $\pm $ {0.085} \\
    \hline    
    \end{tabular}%
    }
\label{tab:addlabel17}%
\end{table*}%

\begin{figure*}[htbp]
    \centering
    \includegraphics[width=\textwidth]{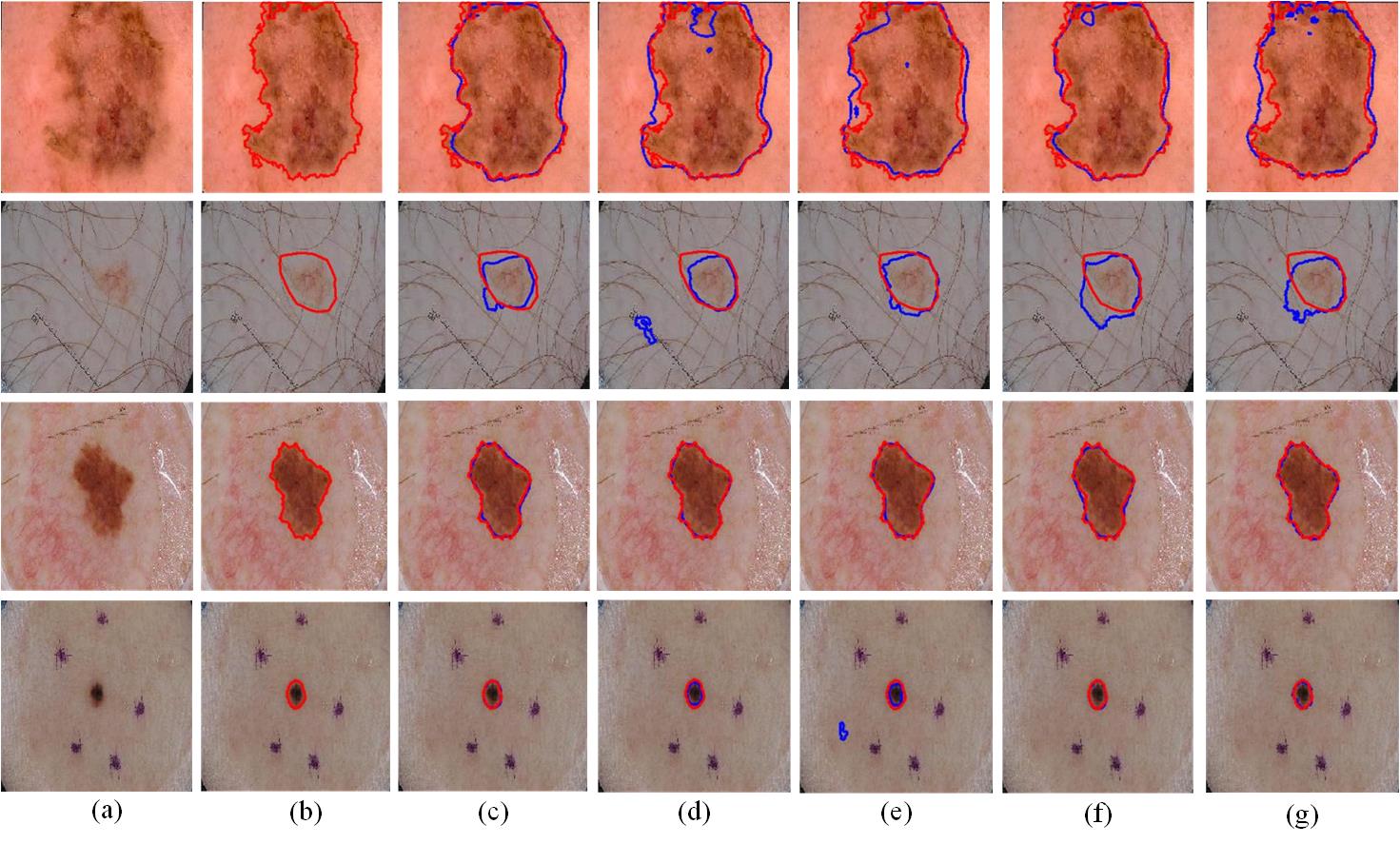}
   \caption{ISIC 2017 dataset visual results: (a) skin lesion images (b) ground truth masks (c) Proposed AD-Net (d) U-Net (e) U-Net++ (f) ARU-GD (g) Swin-Unet. Red contours show the ground truths and blue contours show the segmentation outcomes}
    \label{Fig: ISIC2017attentionmaps}
\end{figure*}

\subsection{\textbf{ISIC 2017 dataset comparison with benchmark models}}

The comparison of AD-Net with 14 SOTA methods on the ISIC 2017 dataset includes U-Net, DAGAN, SUNet-DCP, FAT-Net, TMAHU Net, RMMLP, Ms RED, UNet++, ADF Net, LeaNet, AS-Net, ARU-GD, Swin-Unet and CFF-Net. Table~\ref{tab:addlabel17} compares the results using these SOTA methods. Our AD-Net achieved improvements of 11.65\%, 11.29\%, 11.20\%, 10.43\%, 9.60\%, 7.89\%, 7.58\%, 7.55\%, 7.08\%, 7.07\%, 4.97\%, 4.63\%, 4.60\%, and 4.24\% in terms of Jaccard index compared to these SOTA methods.
{Additionally, in Table~\ref{tab:addlabel17}, we contrast the performance of AD-Net with other prominent methods, including U-Net, Unet++, Swin-Unet, and ARU-GD. We conducted extensive training and analysis of these methods to thoroughly assess the efficiency of AD-Net. Each assessment metric shown in Table~\ref{tab:addlabel18} consists of the mean and standard deviation of the test data set for the U-Net, Unet ++, Swin-Unet, and ARU-GD models. The other methods in Table~\ref{tab:addlabel17} are the articles cited in the skin lesion segmentation domain.}

In addition, visual results were obtained that show various challenges in skin lesion segmentation, such as irregular shapes, hair, and the presence of artifacts. Figure~\ref{Fig: ISIC2017attentionmaps} presents these visual results, illustrating AD-Net's superior performance on unseen test data.


\begin{table*}[htbp]
  \centering
  \caption{{Comparing the proposed AD-Net's performance (mean $\pm$ standard deviation) against alternative SOTA techniques using the ISIC 2016 dataset. Numbers in bold indicate the best performance.}}
  \adjustbox{max width=\textwidth}{
  \begin{tabular}{lcccccc}
    \hline
    \textbf{Model name} & \textbf{Parameters}  & {\textbf{J}} & {\textbf{D}} & {\textbf{Acc}} & {\textbf{Sn}} & {\textbf{Sp}} \\
    \hline
    DAGAN \cite{LEI2020101716} & 56 &  84.42 & 90.85 & 95.82 & 92.28 & 95.68 \\
    FAT-Net \cite{WU2022102327} & 30 & 85.30  & 91.59  & 96.04 & {92.59} & 96.02 \\
    CFF-Net \cite{qin2023dynamically}  & 9.71 & 85.71 & 92.12  & --- & 90.71 & ----\\
    H2Former \cite{he2023h2former} & 33.71 & 86.45 & 92.41 & 96.31 & --- & --- \\
    Ms RED \cite{DAI2022102293} & 3.80 & 87.03 & 92.66 & 96.42 & --- & --- \\
    {ADF-Net} \cite{huang2024adf} & 36.21 & 87.40 & 92.89 & 96.53 & 94.45 & 96.41 \\
   {TMAHU-Net} \cite{dong2024transformer} & 37.43 & 88.19 & 93.73  & --- & 93.55 & --- \\
    \hdashline
    U-Net \cite{ronneberger2015u} & 33.10 & 81.38 $\pm$ {0.127} & 88.24 $\pm$ {0.104} & 93.31 $\pm$ {0.060} & 87.28 $\pm$ {0.086} & 92.88 $\pm$ {0.104}\\  
    UNet++ \cite{zhou2018unet++} & 9.20 & 82.81 $\pm$ {0.118}  & 89.19 $\pm$ {0.093} & 93.88 $\pm$ {0.051} & 88.78 $\pm$ {0.075} & 93.52 $\pm$ {0.089}\\
    ARU-GD \cite{maji2022attention} & 33.50 & 85.12 $\pm$ {0.085} & 90.83 $\pm$ {0.064} & 94.38 $\pm$ {0.048} & 89.86 $\pm$ {0.053} & 94.65 $\pm$ {0.069} \\
    Swin-Unet \cite{cao2023swin} & 2.07  & 85.77 $\pm$ {0.150} &  91.43 $\pm$ {0.126} & 95.52 $\pm$ {0.068} & 93.37 $\pm ${0.086} & 94.48 $\pm $ {0.140}\\
    \hline
    {Proposed} &  \textbf{2.92} & \textbf{89.91} $\pm $ {0.099} & \textbf{94.30} $\pm$ {0.076} & \textbf{97.10}$\pm $ {0.061}& \textbf{94.43} $\pm $ {.096} & \textbf{96.67} $\pm$ {0.052} \\
      \hline
    \end{tabular}%
    }
  \label{tab:addlabel16}%
\end{table*}%

\begin{figure*}[htbp]
    \centering
    \includegraphics[width=\textwidth]{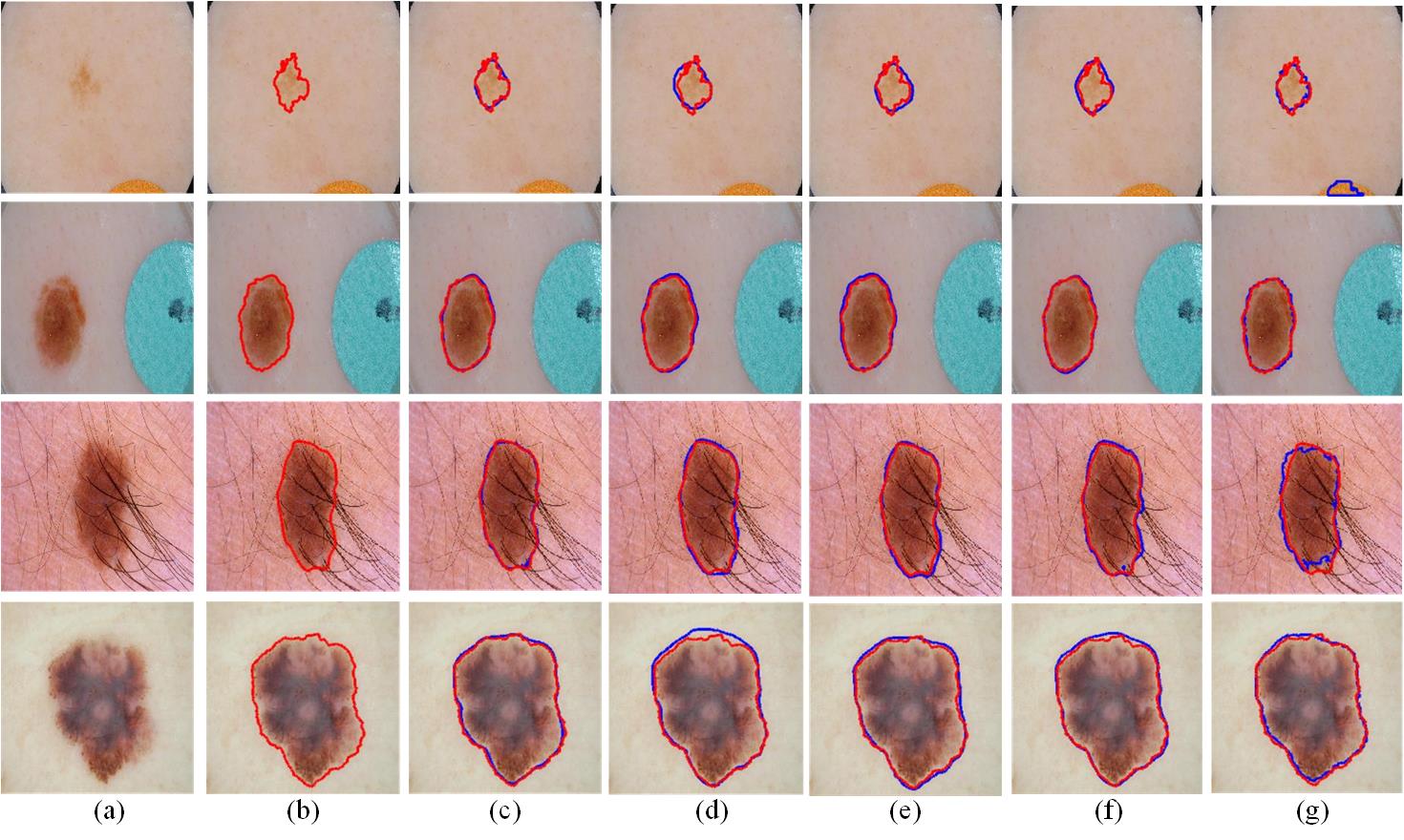}
    \caption{ISIC 2016 dataset visual results: (a) skin lesion images (b) ground truth masks (c) Proposed AD-Net (d) U-Net (e) U-Net++ (f) ARU-GD (g) Swin-Unet. Red contours show the ground truths and blue contours show the segmentation outcomes}
    \label{Fig: ISIC2016attentionmaps}
\end{figure*}

\subsection{\textbf{ISIC 2016 dataset comparison with benchmark models}}
The proposed AD-Net was compared with eleven SOTA methods on the ISIC 2016 dataset, including U-Net, UNet++, DAGAN, ARU-GD, FAT-Net, CFF-Net, Swin-Unet, H2Former, Ms RED, ADF Net, and TMAHU Net. Table~\ref{tab:addlabel16} compares the results with these methods.
Our method demonstrated improvements of 10.48\%, 8.57\%, 6.50\%, 5.63\%, 5.40\%, 4.90\%, 4.83\%, 4.00\%, 3.31\%, 1.73\%, and 0.82\% in terms of the Jaccard index compared to these cutting-edge approaches.
{olor In Table~\ref{tab:addlabel16}, we contrast the performance of AD-Net with the cited papers in skin lesion segmentation and with key SOTA methods such as U-Net, Unet++, Swin-Unet and ARU-GD. Comprehensive training and analysis of U-Net, Unet++, Swin-Unet, and ARU-GD were performed to provide a thorough comparison of efficiency against AD-Net. Each assessment metric shown in Table~\ref{tab:addlabel18} consists of the mean and standard deviation of the test data set for the U-Net, Unet++, Swin-Unet, and ARU-GD models.}

In addition, visual results were obtained that show various challenges encountered in the segmentation of skin lesions, including irregular shapes, hair, artifacts, and multiple lesions. Figure~\ref{Fig: ISIC2016attentionmaps} illustrates these challenges, demonstrating AD-Net's ability to achieve state-of-the-art performance in the segmentation of skin lesions on unseen test data. This demonstrates the stability and efficacy of our method in precisely identifying skin lesions in a variety of settings.

\subsection{\textbf{PH2 dataset comparison with benchmark models}}

Using the PH2 dataset, we analyzed our AD-Net compared to seven cutting-edge methods to compare their performance. 
The goal was to evaluate AD-Net's generalisability and effectiveness against other techniques. Among these approaches are multistage FCN, FCN+BPB+SBE, DCL-PSI, RMMLP, T-Net, ICL-Net, and AS-Net. For this evaluation, we trained our proposed method on the ISIC 2016 dataset and evaluated its performance on the PH2 dataset as a test set. Table~\ref{tab:addlabelph2} presents the results of this comparison. In terms of the Jaccard index, our proposed AD-Net showed improvements of 5.33\%, 4.95\%, 2.99\%, 2.49\%, 1.74\%, 1.40\%, and 1.0\% compared to the respective state-of-the-art techniques.

\begin{figure*}[htbp]
    \centering
    \includegraphics[width=\textwidth]{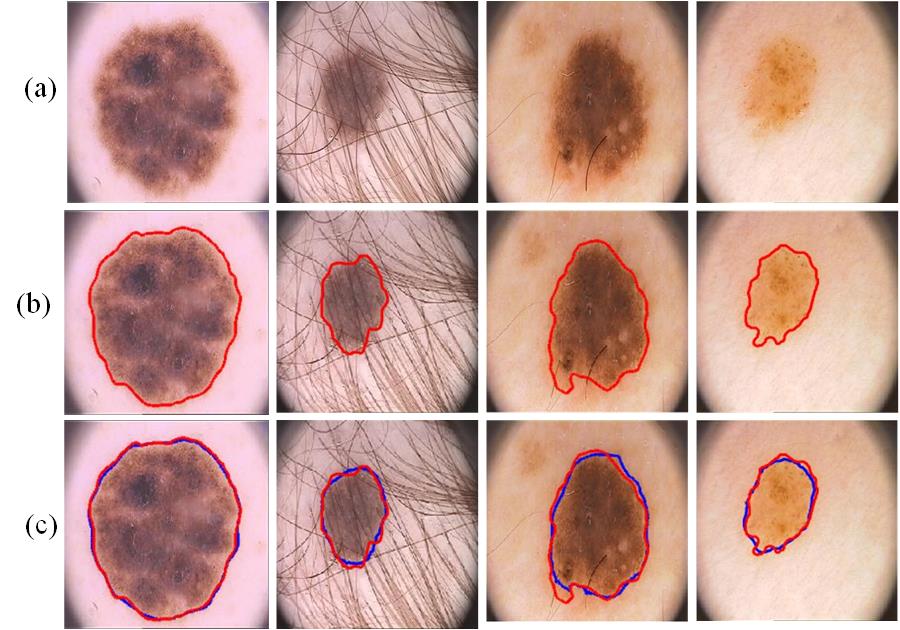}
    \caption{PH2 dataset visual results: (a) skin lesion images (b) ground truth masks (c) Proposed AD-Net. Red contours show the ground truths and blue contours show the segmentation outcomes}
    \label{Fig: ph2attentionmaps}
    \vspace{0.5cm}
\end{figure*}

Figure~\ref{Fig: ph2attentionmaps} provides visual representations of various challenges in the segmentation of skin lesions encountered during this evaluation. This evaluation demonstrates that our AD-Net achieves significant performance gains over existing methods when applied to the PH2 dataset, showcasing its robustness and capability in managing various skin lesion challenges effectively.

\begin{table*}[htbp]
  \centering
  \caption{{
The contrast experiment results, presented as mean $\pm$ standard deviation, show the outcomes of training on the ISIC 2016 dataset and testing on the PH2 dataset.}}
  \adjustbox{max width=\textwidth}{
  \begin{tabular}{lccccc}
    \hline   
    \textbf{Model name}   & {\textbf{J}} & {\textbf{D}} & {\textbf{Acc}} & {\textbf{Sn}} & {\textbf{Sp}} \\
   \hline
    Multistage FCN \cite{bi2017dermoscopic}   & 83.99 & 90.66 & 94.24 & 94.89 & 93.98 \\
    FCN+BPB+SBE \cite{lee2020structure}   & 84.30 & 91.84 & --- & --- & --- \\
    DCL-PSI \cite{bi2019step}   & 85.90 & 92.10 & 95.30 & 96.23 & 94.52 \\
    RMMLP \cite{ji2023rmmlp}    & 86.32  & 92.62 & --- & --- &  ---\\
    T-Net \cite{khan2022t}      & 86.96  & 92.82 & --- & --- &  ---\\
    ICL-Net \cite{cao2022icl}   & 87.25 & 92.80 & \textbf{96.32} & 95.46 & \textbf{97.36} \\
    AS-Net \cite{HU2022117112}  & 87.60 & 93.05 & 95.20 & \textbf{96.24 } & 94.31 \\
    \hline
    {Proposed} &  \textbf{88.47} $\pm $ {0.098}   & \textbf{93.55} $\pm $ {0.064} & 96.19 $\pm $ {0.046}  & {94.89} $\pm $ {0.111}  & 94.41 $\pm $ {0.041}  \\
    \hline
    \end{tabular}%
    }
  \label{tab:addlabelph2}%
\end{table*}

\section{Discussion}
In this section, we show the outcomes of the proposed AD-Net, including its performance at different image resolutions, computational considerations, statistical analysis, and limitations.
\subsection{Outcomes on different number of image resolutions}
{A detailed comparative analysis of the proposed AD-Net's performance across different image resolutions is presented in Table~\ref{tab:multiresolution18} for the ISIC 2018 dataset. The nearest-neighbour algorithm is used to reduce the resolution of the image. This evaluation focuses on three specific image resolutions: $512 \times 512$, $256 \times 256$, and $224 \times 224$. The results show that using an image size of $512 \times 512$ slightly produces the highest performance in the Jaccard index, the dice coefficient, and the sensitivity compared to the two resolutions.
This improvement at the 512$\times$512 resolution can be attributed to the improved ability of the network to capture finer details and subtle variations in skin lesions, which are critical for accurate segmentation. Higher resolution allows the model to leverage more detailed spatial information, leading to better feature extraction and more precise decision-making.
In contrast, reduced resolutions of $256 \times 256$ and $224 \times 224$ can lead to a loss of important structural and textural information, thus slightly reducing the overall efficacy of the network. The outcomes of the ISIC 2018 dataset suggest that, for applications where computational resources allow, opting for higher-resolution images can be beneficial to improve the accuracy and reliability of automated skin lesion segmentation. However, the trade-off between computational load and performance gains should be carefully considered, especially in resource-constrained settings.}

\begin{table*}[htbp]
  \centering
  \caption{{Results of the AD-Net using various image resolutions on the ISIC 2018 dataset}}
    \adjustbox{max width=\textwidth}{
    \begin{tabular}{lcccccc}
    \hline
     \textbf{Model name} & \textbf{Image resolution}  & {\textbf{J}} & {\textbf{D}} & {\textbf{Acc}} & {\textbf{Sn}} & {\textbf{Sp}} \\
    \hline  
    
    Proposed method & $224\times224$  & 86.81 & 92.17 & 95.36 & 93.16 & 94.05 \\
    Proposed method & $256\times256$ & 87.39 & 92.53 & 95.64 & 93.15 & 94.92 \\
    Proposed method & $512\times512$ & 87.73 & 92.73 & 94.87 & 93.30  & 93.98\ \\
    \hline
    \end{tabular}%
    }
  \label{tab:multiresolution18}%
\end{table*}%

{Table~\ref{tab:multiresolution17} presents a comparison of the proposed AD-Net's performance across different image resolutions for the ISIC 2017 dataset. The results indicate that the $512 \times 512$ image size achieves slightly superior results except for specificity compared to the $256 \times 256$ and $224 \times 224$ image sizes. The proposed AD-Net achieves the highest specificity on $224 \times 224$ image size. }
\begin{table*}[htbp]
  \centering
  \caption{{Results of the AD-Net using various image resolutions on the ISIC 2017 dataset}}
    \adjustbox{max width=\textwidth}{
    \begin{tabular}{lcccccc}
    \hline
    \textbf{Model name} & \textbf{Image resolution}  & {\textbf{J}} & {\textbf{D}} & {\textbf{Acc}} & {\textbf{Sn}} & {\textbf{Sp}} \\
    \hline  
    Proposed method & $224\times224$  & 83.36 & 89.84 & 95.18 & 89.65 & 97.20 \\
    Proposed method & $256\times256$ & 84.51 & 90.86 & 95.82 & 90.22 & 96.47 \\
    Proposed method & $512\times512$  & 84.96 & 91.17 & 95.93 & 90.47  & 96.59\\
  
    \hline
    \end{tabular}%
    }
  \label{tab:multiresolution17}%
\end{table*}%

{Table~\ref{tab:multiresolution16} compares the proposed AD-Net's performance across different image resolutions for the ISIC 2016 dataset. The evaluation encompasses three specific resolutions: $512 \times 512$, $256 \times 256$, and $224 \times 224$. The findings reveal that the $512 \times 512$ image size consistently achieves slightly superior results in terms of key performance metrics such as Jaccard index, dice coefficient, accuracy, and sensitivity. From the Tables~\ref{tab:multiresolution18}, ~\ref{tab:multiresolution17},~\ref{tab:multiresolution16}, It can be seen that the performance is slightly improved in some performance measures, but the computation exponentially increases on higher resolution images.}

\begin{table*}[htbp]
  \centering
  \caption{{Results of the AD-Net using various image resolutions on the ISIC 2016 dataset}}
    \adjustbox{max width=\textwidth}{
    \begin{tabular}{lccccccc}
    \hline
    \textbf{Model name} & \textbf{Image resolution}  & {\textbf{J}} & {\textbf{D}} & {\textbf{Acc}} & {\textbf{Sn}} & {\textbf{Sp}} \\
    \hline  
    
    Proposed method & $224\times224$  &  88.25 & 93.23 & 96.59 & 93.71  & 95.95\\ 
    Proposed method & $256\times256$ & 89.91 & 94.30 & 97.10 & 94.43 & 96.67 \\
    Proposed method & $512\times512$ & 89.95 & 94.26 & 97.15 & 94.79 & 96.62\\
  
    \hline
    \end{tabular}%
    }
  \label{tab:multiresolution16}%
\end{table*}%

\subsection{Computational Analysis}

The computational complexity of the proposed method includes crucial performance metrics such as memory size, inference time, parameter count, and floating-point operations (FLOPs). These metrics provide valuable insights into the method's resource requirements and computational efficiency compared to alternative approaches. This analysis facilitates the evaluation of whether the proposed approach achieves an optimal balance between computational efficiency and performance \cite{bianco2018benchmark, iqbal2023ldmres}.

A summary of computational complexity is presented in Table~\ref{computational}. It demonstrates that the proposed AD-Net outperforms other state-of-the-art methods in terms of parameters, FLOPs, inference time, and memory size. These findings establish the proposed AD-Net as a preferred choice for clinical applications, highlighting its efficiency and suitability for practical deployment.


\begin{table*}[htbp]
  \centering
  \caption{{Analysis of AD-Net's computational complexity using the ISIC 2017 dataset. The bolded results are the ones that stand out as performing better than the others that are displayed.
}}
  \adjustbox{max width=\textwidth}{
  \begin{tabular}{lccccc}
    \hline
   \textbf{Model name} &\textbf{Parameters (M) } & \textbf{FLOPs (G) } & \textbf{Inference Time (ms)} & \textbf{Size (MB)} & \textbf{J} \\
   \hline
   U-Net \cite{ronneberger2015u} & 31.12 & 169.35 & 42.1 & 121.57 &75.69\\
   FAT-Net \cite{WU2022102327}& 30 & 23 & --& --& 76.53\\
   UNet++ \cite{zhou2018unet++} & 9.1 & 59.6 & 35.5 & 34.49 & 78.58\\
   SEACU-Net \cite{jiang2022seacu} & 12.81 & -- & 36 &--& 80.50\\
   ARU-GD \cite{maji2022attention} & 33.5 & 104 & 35.8 & 127.82 & 80.77\\
   \hline
   Proposed & \textbf{2.92} & \textbf{4.16} & \textbf{11.42} & \textbf{10.28} & \textbf{84.51}\\
    \hline
    \end{tabular}%
    }
  \label{computational}%
\end{table*}

\subsection{Statistical analysis}
{Based on the results of the Paired Wilcoxon signed-rank test (at the 5\% significance level), significant improvements were observed with the proposed AD-Net compared to ARU-GD \cite{maji2022attention} across different test datasets. Specifically, on the ISIC 2017 dataset, the Jaccard index showed a substantial enhancement with a value of $p$ less than 1.82e-24.
Similarly, for the ISIC 2016, ISIC 2018, and PH2 datasets, the $p$ values were 1.69e-38, 1.52e-42, and 2.92e-7, respectively. These findings underscore the significant performance superiority of AD-Net over ARU-GD in skin lesion segmentation tasks.}


\subsection{Limitation}
\begin{figure*}[htbp]
    \centering
    \includegraphics[scale=0.53]{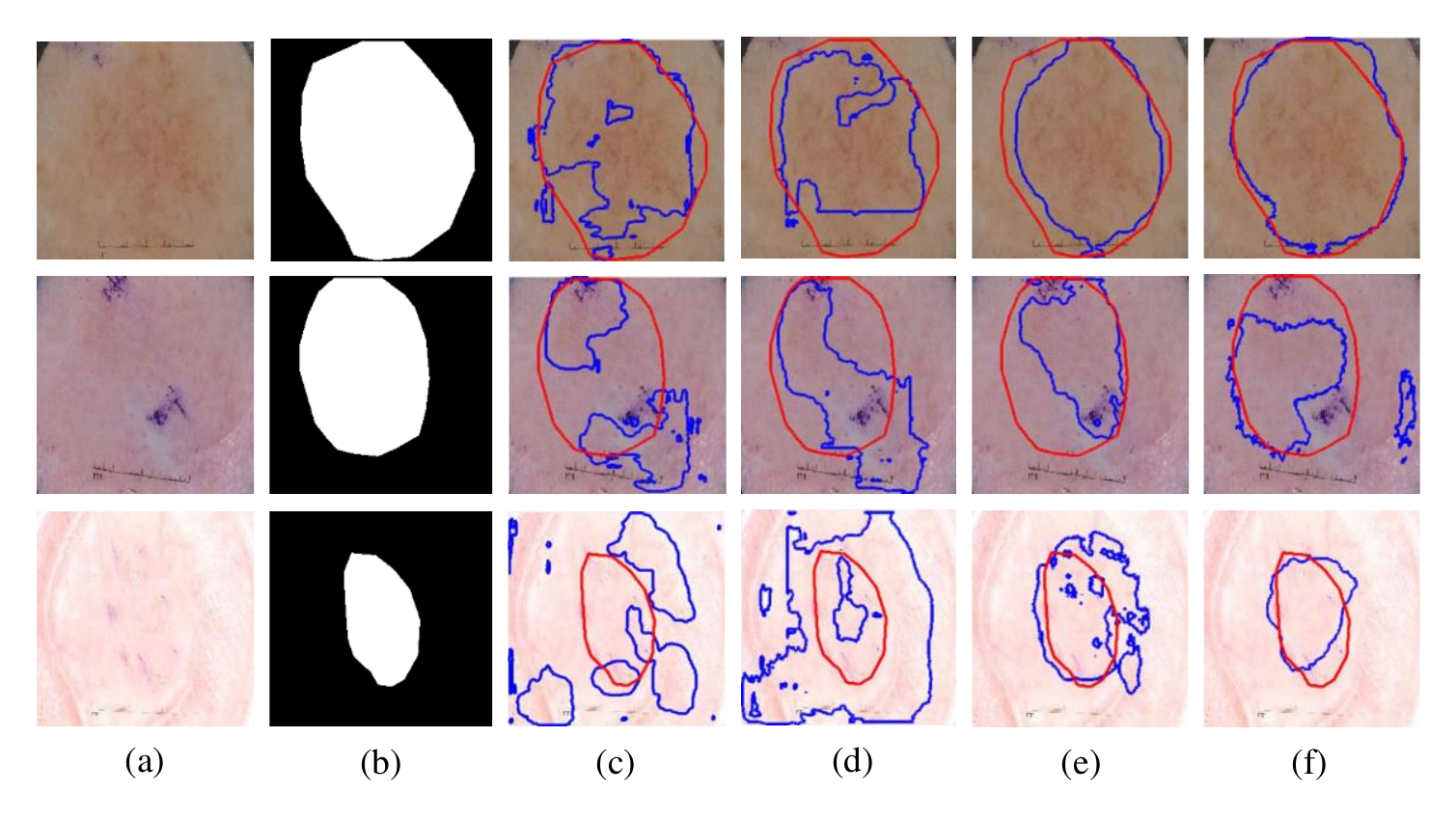}
    \caption{{Limitation case's visual results (a) skin lesion images (b) corresponding ground truth masks (c) U-Net \cite{ronneberger2015u} (d) U-Net++ \cite{zhou2018unet++} (e) ARU-GD \cite{maji2022attention} (f) proposed AD-Net. Red contours indicate the ground truth and blue contours show the segmentation outcomes of different techniques.}}
    \label{Fig: limitation}
    \vspace{0.5cm}
\end{figure*}

{
While the proposed AD-Net generally outperforms existing state-of-the-art techniques, there are specific scenarios in which its performance may be limited. This limitation is particularly evident in images with low contrast between the lesions and the surrounding healthy tissue. As illustrated in Figure\ref{Fig: limitation}, accurately delineating the borders of skin lesions becomes a challenge for AD-Net and other techniques under such conditions. Despite these challenges, AD-Net demonstrates superior segmentation efficiency compared to its competitors. This highlights AD-Net as a significant advancement in skin lesion segmentation, delivering improved outcomes even in challenging scenarios.}

\section{Conclusion and Future Work}

\subsection{Conclusion}
To address challenges in skin lesion segmentation, a novel method has been introduced. The proposed method consists of an attention-based spatial feature enhancement block (ASFEB), dilated convolutional residual blocks with different receptive fields, and a guided decoder strategy. The ASFEB plays a crucial role in skip connections in improving feature fusion, attention weighting, and spatial information, thus improving the model's ability to handle variations in skin lesions. The guided decoder strategy facilitates the fast gradient flow and refines the feature information. The authors also consider the effect of trainable parameters for the skin lesion segmentation task. The proposed AD-Net has shown superior performance compared to several SOTA approaches. The evaluation was carried out on four publicly available benchmark datasets for the segmentation of skin lesions.

\subsection{Future Work}

{Future research will focus on the development and refinement of innovative deep-learning architectures and algorithms, particularly generative models. These advanced models have the potential to enhance segmentation performance by effectively handling low-contrast scenarios and other challenging conditions.
Our aim is to design architectures that can better differentiate between lesions and surrounding healthy tissue in low-contrast images. By using generative networks to create high-fidelity synthetic data, we can augment training datasets, leading to improved model performance and robustness. Additionally, we will combine information from multiple scales and modalities to enhance the segmentation of lesions with various appearances and characteristics.}








\section*{Conflict of Interest} 
The authors have declared no conflicts of interest.

\backmatter

\section*{Data Availability}
The ISIC datasets are accessible to the general public through the following links: \href{https://challenge.isic-archive.com/data/}{ISIC archive}. Additionally open to the public, the PH2 \cite{mendoncca2013ph} dataset can be obtained via \href {https://www.fc.up.pt/addi/ph2%20database.html}{PH2}.


\section*{Funding}

There was no outside funding for this task.

\section*{Ethics approval} 
The authors certify that no research using humans or animals was done for this paper.

\bibliography{references}

\begin{thebibliography}{90}
\providecommand{\natexlab}[1]{#1}
\providecommand{\url}[1]{{#1}}
\providecommand{\urlprefix}{URL }
\providecommand{\doi}[1]{\url{https://doi.org/#1}}
\providecommand{\eprint}[2][]{\url{#2}}
 \bibcommenthead

\bibitem[{Abdullah et~al(2021)Abdullah, Imtiaz, Madni, Khan, Khan, Khan, and
  Naqvi}]{abdullah2021review}
Abdullah F, Imtiaz R, Madni HA, et~al (2021) A review on glaucoma disease
  detection using computerized techniques. IEEE Access 9:37311--37333

\bibitem[{Alahmadi(2022)}]{alahmadi2022multiscale}
Alahmadi MD (2022) Multiscale attention u-net for skin lesion segmentation.
  IEEE Access 10:59145--59154

\bibitem[{Alom et~al(2019)Alom, Yakopcic, Hasan, Taha, and
  Asari}]{alom2019recurrent}
Alom MZ, Yakopcic C, Hasan M, et~al (2019) Recurrent residual u-net for medical
  image segmentation. Journal of Medical Imaging 6(1):014006--014006

\bibitem[{Altan(2021)}]{altan2021enhancing}
Altan G (2021) Enhancing deep learning-based organ segmentation for diagnostic
  support systems on chest x-rays. In: Deep Learning for Biomedical
  Applications. CRC Press, p 255--267

\bibitem[{Altan(2022)}]{altan2022deepoct}
Altan G (2022) Deepoct: an explainable deep learning architecture to analyze
  macular edema on oct images. Engineering Science and Technology, an
  International Journal 34:101091

\bibitem[{Arsalan et~al(2022)Arsalan, Khan, Naqvi, Nawaz, and
  Razzak}]{arsalan2022prompt}
Arsalan M, Khan TM, Naqvi SS, et~al (2022) Prompt deep light-weight vessel
  segmentation network (plvs-net). IEEE/ACM Transactions on Computational
  Biology and Bioinformatics 20(2):1363--1371

\bibitem[{Azad et~al(2019)Azad, Asadi-Aghbolaghi, Fathy, and
  Escalera}]{azad2019bi}
Azad R, Asadi-Aghbolaghi M, Fathy M, et~al (2019) Bi-directional convlstm u-net
  with densley connected convolutions. In: Proceedings of the IEEE/CVF
  international conference on computer vision workshops, pp 0--0

\bibitem[{Balch et~al(2009)Balch, Gershenwald, Soong, Thompson, Atkins, Byrd,
  Buzaid, Cochran, Coit, Ding et~al}]{balch2009final}
Balch CM, Gershenwald JE, Soong Sj, et~al (2009) Final version of 2009 ajcc
  melanoma staging and classification. Journal of clinical oncology 27(36):6199

\bibitem[{van Beers et~al(2019)van Beers, Lindstr{\"o}m, Okafor, and
  Wiering}]{van2019deep}
van Beers F, Lindstr{\"o}m A, Okafor E, et~al (2019) Deep neural networks with
  intersection over union loss for binary image segmentation. In: ICPRAM, pp
  438--445

\bibitem[{Bi et~al(2017)Bi, Kim, Ahn, Kumar, Fulham, and
  Feng}]{bi2017dermoscopic}
Bi L, Kim J, Ahn E, et~al (2017) Dermoscopic image segmentation via multistage
  fully convolutional networks. IEEE Transactions on Biomedical Engineering
  64(9):2065--2074

\bibitem[{Bi et~al(2019)Bi, Kim, Ahn, Kumar, Feng, and Fulham}]{bi2019step}
Bi L, Kim J, Ahn E, et~al (2019) Step-wise integration of deep class-specific
  learning for dermoscopic image segmentation. Pattern recognition 85:78--89

\bibitem[{Bianco et~al(2018)Bianco, Cadene, Celona, and
  Napoletano}]{bianco2018benchmark}
Bianco S, Cadene R, Celona L, et~al (2018) Benchmark analysis of representative
  deep neural network architectures. IEEE access 6:64270--64277

\bibitem[{Cao et~al(2023)Cao, Wang, Chen, Jiang, Zhang, Tian, and
  Wang}]{cao2023swin}
Cao H, Wang Y, Chen J, et~al (2023) Swin-unet: Unet-like pure transformer for
  medical image segmentation. In: Computer Vision--ECCV 2022 Workshops: Tel
  Aviv, Israel, October 23--27, 2022, Proceedings, Part III, Springer, pp
  205--218

\bibitem[{Cao et~al(2022)Cao, Yuan, Liu, Peng, Xie, Yang, Ni, and
  Zheng}]{cao2022icl}
Cao W, Yuan G, Liu Q, et~al (2022) Icl-net: Global and local inter-pixel
  correlations learning network for skin lesion segmentation. IEEE Journal of
  Biomedical and Health Informatics

\bibitem[{Chen et~al(2021)Chen, Liu, Zhang, Lu, and
  Zhang}]{chen2021transattunet}
Chen B, Liu Y, Zhang Z, et~al (2021) Transattunet: Multi-level attention-guided
  u-net with transformer for medical image segmentation. arXiv preprint
  arXiv:210705274

\bibitem[{Codella et~al(2019)Codella, Rotemberg, Tschandl, Celebi, Dusza,
  Gutman, Helba, Kalloo, Liopyris, Marchetti et~al}]{codella2019skin}
Codella N, Rotemberg V, Tschandl P, et~al (2019) Skin lesion analysis toward
  melanoma detection 2018: A challenge hosted by the international skin imaging
  collaboration (isic). arXiv preprint arXiv:190203368

\bibitem[{Codella et~al(2018)Codella, Gutman, Celebi, Helba, Marchetti, Dusza,
  Kalloo, Liopyris, Mishra, Kittler et~al}]{codella2018skin}
Codella NC, Gutman D, Celebi ME, et~al (2018) Skin lesion analysis toward
  melanoma detection: A challenge at the 2017 international symposium on
  biomedical imaging (isbi), hosted by the international skin imaging
  collaboration (isic). In: 2018 IEEE 15th international symposium on
  biomedical imaging (ISBI 2018), IEEE, pp 168--172

\bibitem[{Dai et~al(2022)Dai, Dong, Xu, Yan, Li, Zhang, and
  Luo}]{DAI2022102293}
Dai D, Dong C, Xu S, et~al (2022) Ms red: A novel multi-scale residual encoding
  and decoding network for skin lesion segmentation. Medical Image Analysis
  75:102293. \doi{https://doi.org/10.1016/j.media.2021.102293}

\bibitem[{Dixit et~al(2021)Dixit, Chaurasia, and Mishra}]{dixit2021dilated}
Dixit M, Chaurasia K, Mishra VK (2021) Dilated-resunet: A novel deep learning
  architecture for building extraction from medium resolution multi-spectral
  satellite imagery. Expert Systems with Applications 184:115530

\bibitem[{Dong et~al(2023)Dong, Dai, Zhang, Zhang, Li, and
  Xu}]{dong2023learning}
Dong C, Dai D, Zhang Y, et~al (2023) Learning from dermoscopic images in
  association with clinical metadata for skin lesion segmentation and
  classification. Computers in Biology and Medicine 152:106321

\bibitem[{Dong et~al(2024)Dong, Li, and Hua}]{dong2024transformer}
Dong Z, Li J, Hua Z (2024) Transformer-based multi-attention hybrid networks
  for skin lesion segmentation. Expert Systems with Applications 244:123016

\bibitem[{Farooq et~al(2024)Farooq, Zafar, Saadat, Khan, Iqbal, and
  Razzak}]{farooq2024lssf}
Farooq H, Zafar Z, Saadat A, et~al (2024) Lssf-net: Lightweight segmentation
  with self-awareness, spatial attention, and focal modulation. arXiv preprint
  arXiv:240901572

\bibitem[{Feng et~al(2022)Feng, Ren, Wang, Wang, and Li}]{FENG2022105942}
Feng K, Ren L, Wang G, et~al (2022) Slt-net: A codec network for skin lesion
  segmentation. Computers in Biology and Medicine 148:105942.
  \doi{https://doi.org/10.1016/j.compbiomed.2022.105942}

\bibitem[{Feng et~al(2020{\natexlab{a}})Feng, Zhao, Shi, Cheng, Wang, Ma,
  Xiang, Zhu, and Chen}]{feng2020cpfnet}
Feng S, Zhao H, Shi F, et~al (2020{\natexlab{a}}) Cpfnet: Context pyramid
  fusion network for medical image segmentation. IEEE transactions on medical
  imaging 39(10):3008--3018

\bibitem[{Feng et~al(2020{\natexlab{b}})Feng, Zhao, Shi, Cheng, Wang, Ma,
  Xiang, Zhu, and Chen}]{9049412}
Feng S, Zhao H, Shi F, et~al (2020{\natexlab{b}}) Cpfnet: Context pyramid
  fusion network for medical image segmentation. IEEE Transactions on Medical
  Imaging 39(10):3008--3018. \doi{10.1109/TMI.2020.2983721}

\bibitem[{Garcia-Arroyo and Garcia-Zapirain(2019)}]{garcia2019segmentation}
Garcia-Arroyo JL, Garcia-Zapirain B (2019) Segmentation of skin lesions in
  dermoscopy images using fuzzy classification of pixels and histogram
  thresholding. Computer methods and programs in biomedicine 168:11--19

\bibitem[{Gu et~al(2022)Gu, Wang, and Zhang}]{gu2022net}
Gu R, Wang L, Zhang L (2022) De-net: a deep edge network with boundary
  information for automatic skin lesion segmentation. Neurocomputing 468:71--84

\bibitem[{Gutman et~al(2016)Gutman, Codella, Celebi, Helba, Marchetti, Mishra,
  and Halpern}]{gutman2016skin}
Gutman D, Codella NC, Celebi E, et~al (2016) Skin lesion analysis toward
  melanoma detection: A challenge at the international symposium on biomedical
  imaging (isbi) 2016, hosted by the international skin imaging collaboration
  (isic). arXiv preprint arXiv:160501397

\bibitem[{Hafhouf et~al(2022)Hafhouf, Zitouni, Megherbi, and
  Sbaa}]{hafhouf2022improved}
Hafhouf B, Zitouni A, Megherbi AC, et~al (2022) An improved and robust
  encoder--decoder for skin lesion segmentation. Arabian Journal for Science
  and Engineering pp 1--15

\bibitem[{He et~al(2023)He, Wang, Li, Du, Xia, and Fu}]{he2023h2former}
He A, Wang K, Li T, et~al (2023) H2former: An efficient hierarchical hybrid
  transformer for medical image segmentation. IEEE Transactions on Medical
  Imaging

\bibitem[{Holzinger(2022)}]{holzinger2022next}
Holzinger A (2022) The next frontier: Ai we can really trust. In: Machine
  Learning and Principles and Practice of Knowledge Discovery in Databases:
  International Workshops of ECML PKDD 2021, Virtual Event, September 13-17,
  2021, Proceedings, Part I, Springer, pp 427--440

\bibitem[{Hu et~al(2024)Hu, Zhou, Yu, Dai, Wang, Tan, and Sun}]{hu2024leanet}
Hu B, Zhou P, Yu H, et~al (2024) Leanet: Lightweight u-shaped architecture for
  high-performance skin cancer image segmentation. Computers in Biology and
  Medicine 169:107919

\bibitem[{Hu et~al(2022)Hu, Lu, Lee, Xiong, and Chen}]{HU2022117112}
Hu K, Lu J, Lee D, et~al (2022) As-net: Attention synergy network for skin
  lesion segmentation. Expert Systems with Applications 201:117112.
  \doi{https://doi.org/10.1016/j.eswa.2022.117112}

\bibitem[{Huang et~al(2024)Huang, Deng, Yin, Zhang, Tang, and
  Wang}]{huang2024adf}
Huang Z, Deng H, Yin S, et~al (2024) Adf-net: A novel adaptive dual-stream
  encoding and focal attention decoding network for skin lesion segmentation.
  Biomedical Signal Processing and Control 91:105895

\bibitem[{Imtiaz et~al(2021)Imtiaz, Khan, Naqvi, Arsalan, and
  Nawaz}]{imtiaz2021screening}
Imtiaz R, Khan TM, Naqvi SS, et~al (2021) Screening of glaucoma disease from
  retinal vessel images using semantic segmentation. Computers \& Electrical
  Engineering 91:107036

\bibitem[{Iqbal et~al(2022)Iqbal, Naqvi, Ahmed, Saadat, and Khan}]{iqbal2022g}
Iqbal S, Naqvi S, Ahmed H, et~al (2022) G-net light: A lightweight modified
  google net for retinal vessel segmentation. In: Photonics, MDPI, pp 923--936

\bibitem[{Iqbal et~al(2023)Iqbal, Khan, Naqvi, Naveed, Usman, Khan, and
  Razzak}]{iqbal2023ldmres}
Iqbal S, Khan TM, Naqvi SS, et~al (2023) Ldmres-net: A lightweight neural
  network for efficient medical image segmentation on iot and edge devices.
  IEEE journal of biomedical and health informatics

\bibitem[{Iqbal et~al(2024{\natexlab{a}})Iqbal, Khan, Naqvi, Naveed, and
  Meijering}]{iqbal2024tbconvl}
Iqbal S, Khan TM, Naqvi SS, et~al (2024{\natexlab{a}}) Tbconvl-net: A hybrid
  deep learning architecture for robust medical image segmentation. arXiv
  preprint arXiv:240903367

\bibitem[{Iqbal et~al(2024{\natexlab{b}})Iqbal, Zeeshan, Mehmood, Khan, and
  Razzak}]{iqbal2024tesl}
Iqbal S, Zeeshan M, Mehmood M, et~al (2024{\natexlab{b}}) Tesl-net: A
  transformer-enhanced cnn for accurate skin lesion segmentation. arXiv
  preprint arXiv:240809687

\bibitem[{Jadon(2020)}]{jadon2020survey}
Jadon S (2020) A survey of loss functions for semantic segmentation. In: 2020
  IEEE Conference on Computational Intelligence in Bioinformatics and
  Computational Biology (CIBCB), IEEE, pp 1--7

\bibitem[{Javed et~al(2024)Javed, Khan, Qayyum, Sowmya, and
  Razzak}]{javed2024region}
Javed S, Khan TM, Qayyum A, et~al (2024) Region guided attention network for
  retinal vessel segmentation. arXiv preprint arXiv:240718970

\bibitem[{Ji et~al(2023)Ji, Deng, Ding, Zhou, and Xiao}]{ji2023rmmlp}
Ji C, Deng Z, Ding Y, et~al (2023) Rmmlp: Rolling mlp and matrix decomposition
  for skin lesion segmentation. Biomedical Signal Processing and Control
  84:104825

\bibitem[{Jiang et~al(2022)Jiang, Jiang, Wang, Yu, and Wang}]{jiang2022seacu}
Jiang X, Jiang J, Wang B, et~al (2022) Seacu-net: Attentive convlstm u-net with
  squeeze-and-excitation layer for skin lesion segmentation. Computer Methods
  and Programs in Biomedicine 225:107076

\bibitem[{Khan et~al(2019)Khan, Khan, Soomro, Mir, and Gao}]{khan2019boosting}
Khan MA, Khan TM, Soomro TA, et~al (2019) Boosting sensitivity of a retinal
  vessel segmentation algorithm. Pattern Analysis and Applications 22:583--599

\bibitem[{Khan et~al(2016)Khan, U.~Khan, Kong, and Kittaneh}]{khan2016stopping}
Khan TM, U.~Khan MA, Kong Y, et~al (2016) Stopping criterion for linear
  anisotropic image diffusion: a fingerprint image enhancement case. EURASIP
  Journal on Image and Video Processing 2016:1--20

\bibitem[{Khan et~al(2020{\natexlab{a}})Khan, Naqvi, Arsalan, Khan, Khan, and
  Haider}]{khan2020exploiting}
Khan TM, Naqvi SS, Arsalan M, et~al (2020{\natexlab{a}}) Exploiting residual
  edge information in deep fully convolutional neural networks for retinal
  vessel segmentation. In: 2020 International Joint Conference on Neural
  Networks (IJCNN), IEEE, pp 1--8

\bibitem[{Khan et~al(2020{\natexlab{b}})Khan, Robles-Kelly, and
  Naqvi}]{khan2020semantically}
Khan TM, Robles-Kelly A, Naqvi SS (2020{\natexlab{b}}) A semantically flexible
  feature fusion network for retinal vessel segmentation. In: International
  Conference on Neural Information Processing, Springer, Cham, pp 159--167

\bibitem[{Khan et~al(2021)Khan, Robles-Kelly, Naqvi, and
  Muhammad}]{khan2021residual}
Khan TM, Robles-Kelly A, Naqvi SS, et~al (2021) Residual multiscale full
  convolutional network (rm-fcn) for high resolution semantic segmentation of
  retinal vasculature. In: Structural, Syntactic, and Statistical Pattern
  Recognition: Joint IAPR International Workshops, S+ SSPR 2020, Padua, Italy,
  January 21--22, 2021, Proceedings, Springer Nature, p 324

\bibitem[{Khan et~al(2022{\natexlab{a}})Khan, Arsalan, Robles-Kelly, and
  Meijering}]{khan2022mkis}
Khan TM, Arsalan M, Robles-Kelly A, et~al (2022{\natexlab{a}}) Mkis-net: a
  light-weight multi-kernel network for medical image segmentation. In:
  International Conference on Digital Image Computing: Techniques and
  Applications (DICTA), 10.1109/DICTA56598.2022.10034573, pp 1--8

\bibitem[{Khan et~al(2022{\natexlab{b}})Khan, Khan, Rehman, Naveed, Afridi,
  Naqvi, and Raazak}]{khan2022width}
Khan TM, Khan MA, Rehman NU, et~al (2022{\natexlab{b}}) Width-wise vessel
  bifurcation for improved retinal vessel segmentation. Biomedical Signal
  Processing and Control 71:103169

\bibitem[{Khan et~al(2022{\natexlab{c}})Khan, Naqvi, and
  Meijering}]{khan2022leveraging}
Khan TM, Naqvi SS, Meijering E (2022{\natexlab{c}}) Leveraging image complexity
  in macro-level neural network design for medical image segmentation.
  Scientific Reports 12(1):22286

\bibitem[{Khan et~al(2022{\natexlab{d}})Khan, Naqvi, Robles-Kelly, and
  Meijering}]{khan2022neural}
Khan TM, Naqvi SS, Robles-Kelly A, et~al (2022{\natexlab{d}}) Neural network
  compression by joint sparsity promotion and redundancy reduction. In:
  International Conference on Neural Information Processing, Springer
  International Publishing Cham, pp 612--623

\bibitem[{Khan et~al(2022{\natexlab{e}})Khan, Robles-Kelly, and
  Naqvi}]{khan2022t}
Khan TM, Robles-Kelly A, Naqvi SS (2022{\natexlab{e}}) T-net: A
  resource-constrained tiny convolutional neural network for medical image
  segmentation. In: Proceedings of the IEEE/CVF winter conference on
  applications of computer vision, pp 644--653

\bibitem[{Khan et~al(2023{\natexlab{a}})Khan, Arsalan, Iqbal, Razzak, and
  Meijering}]{khan2023feature}
Khan TM, Arsalan M, Iqbal S, et~al (2023{\natexlab{a}}) Feature enhancer
  segmentation network (fes-net) for vessel segmentation. In: 2023
  International Conference on Digital Image Computing: Techniques and
  Applications (DICTA), IEEE, pp 160--167

\bibitem[{Khan et~al(2023{\natexlab{b}})Khan, Naqvi, Robles-Kelly, and
  Razzak}]{khan2023retinal}
Khan TM, Naqvi SS, Robles-Kelly A, et~al (2023{\natexlab{b}}) Retinal vessel
  segmentation via a multi-resolution contextual network and adversarial
  learning. Neural Networks 165:310--320

\bibitem[{Khan et~al(2024{\natexlab{a}})Khan, Iqbal, Naqvi, Razzak, and
  Meijering}]{khan2024lmbf}
Khan TM, Iqbal S, Naqvi SS, et~al (2024{\natexlab{a}}) Lmbf-net: A lightweight
  multipath bidirectional focal attention network for multifeatures
  segmentation. arXiv preprint arXiv:240702871

\bibitem[{Khan et~al(2024{\natexlab{b}})Khan, Naqvi, and
  Meijering}]{khan2024esdmr}
Khan TM, Naqvi SS, Meijering E (2024{\natexlab{b}}) Esdmr-net: A lightweight
  network with expand-squeeze and dual multiscale residual connections for
  medical image segmentation. Engineering Applications of Artificial
  Intelligence 133:107995

\bibitem[{Kingma and Ba(2014)}]{kingma2014adam}
Kingma DP, Ba J (2014) Adam: A method for stochastic optimization. arXiv
  preprint arXiv:14126980

\bibitem[{Lee et~al(2020)Lee, Kim, Lee, Kim, and Ro}]{lee2020structure}
Lee HJ, Kim JU, Lee S, et~al (2020) Structure boundary preserving segmentation
  for medical image with ambiguous boundary. In: Proceedings of the IEEE/CVF
  Conference on Computer Vision and Pattern Recognition, pp 4817--4826

\bibitem[{Lei et~al(2020)Lei, Xia, Jiang, Jiang, Ge, Xu, Qin, Chen, Wang, and
  Wang}]{LEI2020101716}
Lei B, Xia Z, Jiang F, et~al (2020) Skin lesion segmentation via generative
  adversarial networks with dual discriminators. Medical Image Analysis
  64:101716. \doi{https://doi.org/10.1016/j.media.2020.101716}

\bibitem[{Litjens et~al(2017)Litjens, Kooi, Bejnordi, Setio, Ciompi,
  Ghafoorian, Van Der~Laak, Van~Ginneken, and S{\'a}nchez}]{litjens2017survey}
Litjens G, Kooi T, Bejnordi BE, et~al (2017) A survey on deep learning in
  medical image analysis. Medical image analysis 42:60--88

\bibitem[{Ma and Tavares(2016)}]{7004778}
Ma Z, Tavares JMRS (2016) A novel approach to segment skin lesions in
  dermoscopic images based on a deformable model. IEEE Journal of Biomedical
  and Health Informatics 20(2):615--623. \doi{10.1109/JBHI.2015.2390032}

\bibitem[{Maji et~al(2022)Maji, Sigedar, and Singh}]{maji2022attention}
Maji D, Sigedar P, Singh M (2022) Attention res-unet with guided decoder for
  semantic segmentation of brain tumors. Biomedical Signal Processing and
  Control 71:103077

\bibitem[{Maqsood and Dama{\v{s}}evi{\v{c}}ius(2023)}]{maqsood2023multiclass}
Maqsood S, Dama{\v{s}}evi{\v{c}}ius R (2023) Multiclass skin lesion
  localization and classification using deep learning based features fusion and
  selection framework for smart healthcare. Neural networks 160:238--258

\bibitem[{Mendon{\c{c}}a et~al(2013)Mendon{\c{c}}a, Ferreira, Marques, Marcal,
  and Rozeira}]{mendoncca2013ph}
Mendon{\c{c}}a T, Ferreira PM, Marques JS, et~al (2013) Ph 2-a dermoscopic
  image database for research and benchmarking. In: 2013 35th annual
  international conference of the IEEE engineering in medicine and biology
  society (EMBC), IEEE, pp 5437--5440

\bibitem[{Naqvi et~al(2019)Naqvi, Fatima, Khan, Rehman, and
  Khan}]{naqvi2019automatic}
Naqvi SS, Fatima N, Khan TM, et~al (2019) Automatic optic disk detection and
  segmentation by variational active contour estimation in retinal fundus
  images. Signal, Image and Video Processing 13:1191--1198

\bibitem[{Naqvi et~al(2023)Naqvi, Langah, Khan, Khan, Bashir, Razzak, and
  Khan}]{naqvi2023glan}
Naqvi SS, Langah ZA, Khan HA, et~al (2023) Glan: Gan assisted lightweight
  attention network for biomedical imaging based diagnostics. Cognitive
  Computation 15(3):932--942

\bibitem[{Naveed et~al(2024{\natexlab{a}})Naveed, Naqvi, Iqbal, Razzak, Khan,
  and Khan}]{naveed2024ra}
Naveed A, Naqvi SS, Iqbal S, et~al (2024{\natexlab{a}}) Ra-net: Region-aware
  attention network for skin lesion segmentation. Cognitive Computation pp
  1--18

\bibitem[{Naveed et~al(2024{\natexlab{b}})Naveed, Naqvi, Khan, and
  Razzak}]{naveed2024pca}
Naveed A, Naqvi SS, Khan TM, et~al (2024{\natexlab{b}}) Pca: Progressive
  class-wise attention for skin lesions diagnosis. Engineering Applications of
  Artificial Intelligence 127:107417

\bibitem[{Oktay et~al(2018)Oktay, Schlemper, Folgoc, Lee, Heinrich, Misawa,
  Mori, McDonagh, Hammerla, Kainz et~al}]{oktay2018attention}
Oktay O, Schlemper J, Folgoc LL, et~al (2018) Attention u-net: Learning where
  to look for the pancreas. arXiv preprint arXiv:180403999

\bibitem[{Panayides et~al(2020)Panayides, Amini, Filipovic, Sharma, Tsaftaris,
  Young, Foran, Do, Golemati, Kurc et~al}]{panayides2020ai}
Panayides AS, Amini A, Filipovic ND, et~al (2020) Ai in medical imaging
  informatics: current challenges and future directions. IEEE journal of
  biomedical and health informatics 24(7):1837--1857

\bibitem[{Qin et~al(2023)Qin, Zheng, Zeng, Chen, Zhai, Genovese, Piuri, and
  Scotti}]{qin2023dynamically}
Qin C, Zheng B, Zeng J, et~al (2023) Dynamically aggregating mlps and cnns for
  skin lesion segmentation with geometry regularization. Computer Methods and
  Programs in Biomedicine 238:107601

\bibitem[{Ronneberger et~al(2015)Ronneberger, Fischer, and
  Brox}]{ronneberger2015u}
Ronneberger O, Fischer P, Brox T (2015) U-net: Convolutional networks for
  biomedical image segmentation. In: International Conference on Medical image
  computing and computer-assisted intervention, Springer, pp 234--241

\bibitem[{Schlemper et~al(2019)Schlemper, Oktay, Schaap, Heinrich, Kainz,
  Glocker, and Rueckert}]{SCHLEMPER2019197}
Schlemper J, Oktay O, Schaap M, et~al (2019) Attention gated networks: Learning
  to leverage salient regions in medical images. Medical Image Analysis
  53:197--207. \doi{https://doi.org/10.1016/j.media.2019.01.012},
  \urlprefix\url{https://www.sciencedirect.com/science/article/pii/S1361841518306133}

\bibitem[{Selvaraju et~al(2017)Selvaraju, Cogswell, Das, Vedantam, Parikh, and
  Batra}]{selvaraju2017grad}
Selvaraju RR, Cogswell M, Das A, et~al (2017) Grad-cam: Visual explanations
  from deep networks via gradient-based localization. In: Proceedings of the
  IEEE international conference on computer vision, pp 618--626

\bibitem[{Siegel et~al(2023)Siegel, Miller, Wagle, and
  Jemal}]{siegel2023cancer}
Siegel RL, Miller KD, Wagle NS, et~al (2023) Cancer statistics, 2023. CA: a
  cancer journal for clinicians 73(1):17--48

\bibitem[{Singh et~al(2019)Singh, Abdel-Nasser, Rashwan, Akram, Pandey,
  Lalande, Presles, Romani, and Puig}]{8832175}
Singh VK, Abdel-Nasser M, Rashwan HA, et~al (2019) Fca-net: Adversarial
  learning for skin lesion segmentation based on multi-scale features and
  factorized channel attention. IEEE Access 7:130552--130565.
  \doi{10.1109/ACCESS.2019.2940418}

\bibitem[{Song et~al(2023)Song, Wang, and Wang}]{song2023decoupling}
Song L, Wang H, Wang ZJ (2023) Decoupling multi-task causality for improved
  skin lesion segmentation and classification. Pattern Recognition 133:108995

\bibitem[{Soomro et~al(2016)Soomro, Khan, Gao, Khan, Paul, and
  Mir}]{soomro2016automatic}
Soomro TA, Khan MA, Gao J, et~al (2016) Automatic retinal vessel extraction
  algorithm. In: 2016 International Conference on Digital Image Computing:
  Techniques and Applications (DICTA), IEEE, pp 1--8

\bibitem[{Sufyan et~al(2023)Sufyan, Shokat, and Ashfaq}]{sufyan2023artificial}
Sufyan M, Shokat Z, Ashfaq UA (2023) Artificial intelligence in cancer
  diagnosis and therapy: Current status and future perspective. Computers in
  Biology and Medicine p 107356

\bibitem[{Tschandl et~al(2018)Tschandl, Rosendahl, and
  Kittler}]{tschandl2018ham10000}
Tschandl P, Rosendahl C, Kittler H (2018) The ham10000 dataset, a large
  collection of multi-source dermatoscopic images of common pigmented skin
  lesions. Scientific data 5(1):1--9

\bibitem[{Wang et~al(2023)Wang, Tang, Xiao, Zhou, Fang, and
  Yang}]{wang2023grenet}
Wang J, Tang Y, Xiao Y, et~al (2023) Grenet: Gradually recurrent network with
  curriculum learning for 2-d medical image segmentation. IEEE Transactions on
  Neural Networks and Learning Systems

\bibitem[{Wu et~al(2020)Wu, Pan, Li, Wen, and Qin}]{wu2020automated}
Wu H, Pan J, Li Z, et~al (2020) Automated skin lesion segmentation via an
  adaptive dual attention module. IEEE transactions on medical imaging
  40(1):357--370

\bibitem[{Wu et~al(2022)Wu, Chen, Chen, Wang, Lei, and Wen}]{WU2022102327}
Wu H, Chen S, Chen G, et~al (2022) Fat-net: Feature adaptive transformers for
  automated skin lesion segmentation. Medical Image Analysis 76:102327.
  \doi{https://doi.org/10.1016/j.media.2021.102327}

\bibitem[{Yu and Koltun(2015)}]{yu2015multi}
Yu F, Koltun V (2015) Multi-scale context aggregation by dilated convolutions.
  arXiv preprint arXiv:151107122

\bibitem[{Yu et~al(2017)Yu, Koltun, and Funkhouser}]{yu2017dilated}
Yu F, Koltun V, Funkhouser T (2017) Dilated residual networks. In: Proceedings
  of the IEEE conference on computer vision and pattern recognition, pp
  472--480

\bibitem[{Yuan et~al(2023)Yuan, Li, Wang, and Fang}]{yuan2023lightweight}
Yuan F, Li K, Wang C, et~al (2023) A lightweight network for smoke semantic
  segmentation. Pattern Recognition 137:109289

\bibitem[{Zafar et~al(2020)Zafar, Gilani, Waris, Ahmed, Jamil, Khan, and
  Sohail~Kashif}]{zafar2020skin}
Zafar K, Gilani SO, Waris A, et~al (2020) Skin lesion segmentation from
  dermoscopic images using convolutional neural network. Sensors 20(6):1601

\bibitem[{Zhang et~al(2023)Zhang, Lu, Zhao, Hu, Su, and Yuan}]{zhang2023accpg}
Zhang W, Lu F, Zhao W, et~al (2023) Accpg-net: A skin lesion segmentation
  network with adaptive channel-context-aware pyramid attention and global
  feature fusion. Computers in Biology and Medicine p 106580

\bibitem[{Zhou et~al(2018)Zhou, Rahman~Siddiquee, Tajbakhsh, and
  Liang}]{zhou2018unet++}
Zhou Z, Rahman~Siddiquee MM, Tajbakhsh N, et~al (2018) Unet++: A nested u-net
  architecture for medical image segmentation. In: Deep Learning in Medical
  Image Analysis and Multimodal Learning for Clinical Decision Support: 4th
  International Workshop, DLMIA 2018, and 8th International Workshop, ML-CDS
  2018, Held in Conjunction with MICCAI 2018, Granada, Spain, September 20,
  2018, Proceedings 4, Springer, pp 3--11

\end{thebibliography}

\end{document}